\documentclass{article}

\usepackage{PRIMEarxiv}

\usepackage[utf8]{inputenc} % allow utf-8 input
\usepackage[T1]{fontenc}    % use 8-bit T1 fonts
\usepackage{hyperref}       % hyperlinks
\usepackage{url}            % simple URL typesetting
\usepackage{booktabs}       % professional-quality tables
\usepackage{amsfonts}       % blackboard math symbols
\usepackage{nicefrac}       % compact symbols for 1/2, etc.
\usepackage{microtype}      % microtypography
\usepackage{lipsum}
\usepackage{multirow}
\usepackage{fancyhdr}       % header
\usepackage{graphicx}       % graphics
\graphicspath{{media/}}     % organize your images and other figures under media/ folder

\title{Cursive Caption Text Detection in Videos
}

\author{
  Ali Mirza \\
  Center of Artificial Intelligence\\
  Mantalus \\
  Melbourne, Australia\\
  \texttt{ali.mirza@mantalus.com} \\
   \And
  Imran Siddiqi \\
  Center of Excellence in Artificial Intelligence \\
  Bahria University \\
  Islamabad, Pakistan\\
  \texttt{imran.siddiqi@bahria.edu.pk} \\
}

\begin{document}
\maketitle

% keywords can be removed
\keywords{Text Detection \and Script Identification\and Deep Neural Networks (DNNs)\and Convolutional Neural Networks (CNNs)\and Video Text\and Video Frames Dataset}

\begin{abstract}
Textual content appearing in videos represents an interesting index for semantic retrieval of videos (from archives), generation of alerts (live streams) as well as high level applications like opinion mining and content summarization. One of the key components of such systems is the detection of textual content in video frames and the same makes the subject of our present study. This paper presents a robust technique for detection of textual content appearing in video frames. More specifically we target text in cursive script taking Urdu text as a case study. Detection of textual regions in video frames is carried out by fine-tuning object detectors based on deep convolutional neural networks for the specific case of text detection. Since it is common to have videos with caption text in multiple-scripts, cursive text is distinguished from Latin text using a script-identification module. Finally, detection and script identification are combined in a single end-to-end trainable system. Experiments on a comprehensive dataset of around 11,000 video frames report an F-measure of 0.91.
\end{abstract}

\maketitle
%--------------------------------------------------------------
%--------------------------------------------------------------
%--------------------------------------------------------------
%--------------------------------------------------------------
%--------------------------------------------------------------
%%%%%%%%%%%%%%%%%%%%%%%% Introduction %%%%%%%%%%%%%%%%%%%%%%%%%

\section{Introduction}\label{sec:intro}
In the recent years, there has been a tremendous increase in the amount of digital multimedia data, especially the video content, both in the form of video archives and live streams. According to statistics~\cite{burgess2018youtube}, 300 hours of video is being uploaded every minute on the YouTube. A key factor responsible for this enormous increase is the availability of low-cost smart phones equipped with cameras. With such huge collections of data, there is a need to have efficient as well as effective retrieval techniques allowing users retrieve the desired content. Traditionally, videos are mostly stored with user assigned annotations or keywords which are called tags. When a content is to be searched, a keyword provided as query is matched with these tags to retrieve the relevant content. The assigned tags, naturally, cannot encompass the rich video content leading to a constrained retrieval. A better and more effective strategy is to search within the actual content rather than simply matching the tags i.e. Content based Image or Video Retrieval. CBVR systems have been researched and developed for long a time and allow a smarter way of retrieving the desired content. The term content may refer to the visual content (for example objects or persons in the video), audio content (the spoken keywords for instance) or the textual content (News tickers, anchor names, score cards etc.). Among these, the focus of our current study lies on textual content. More specifically, we target a smart retrieval system that exploits the textual content in videos as an index. \\

\begin{figure}[!ht]
	\centering
		\includegraphics[width=0.95\textwidth]{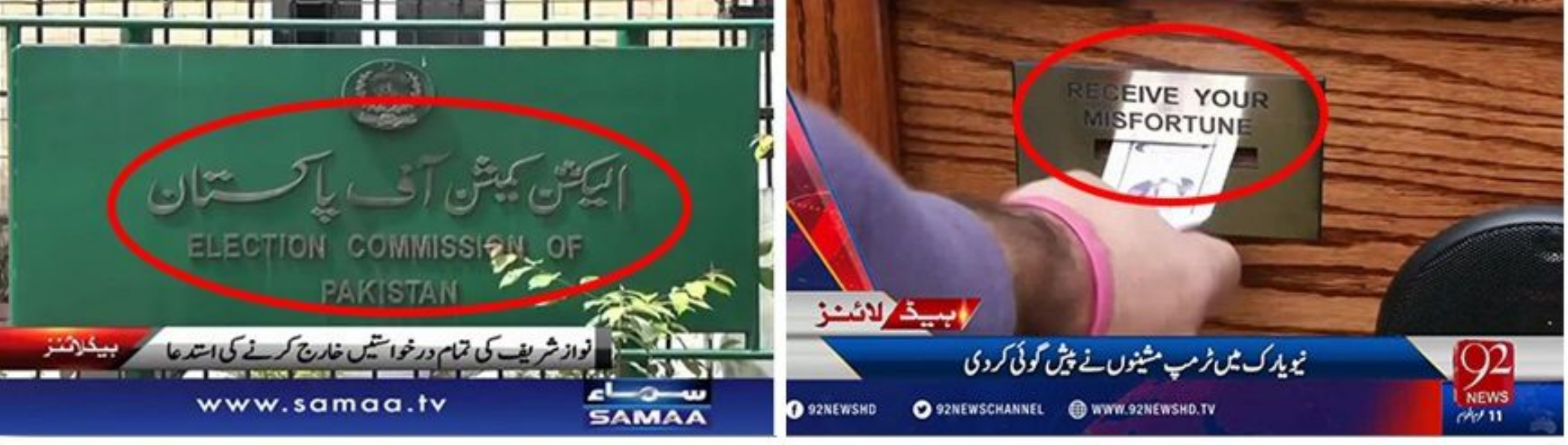}
	\caption{Examples of Scene Text}
		\label{fig:scenetext}
\end{figure}

\begin{figure}[!ht]
	\centering
		\includegraphics[width=0.95\textwidth]{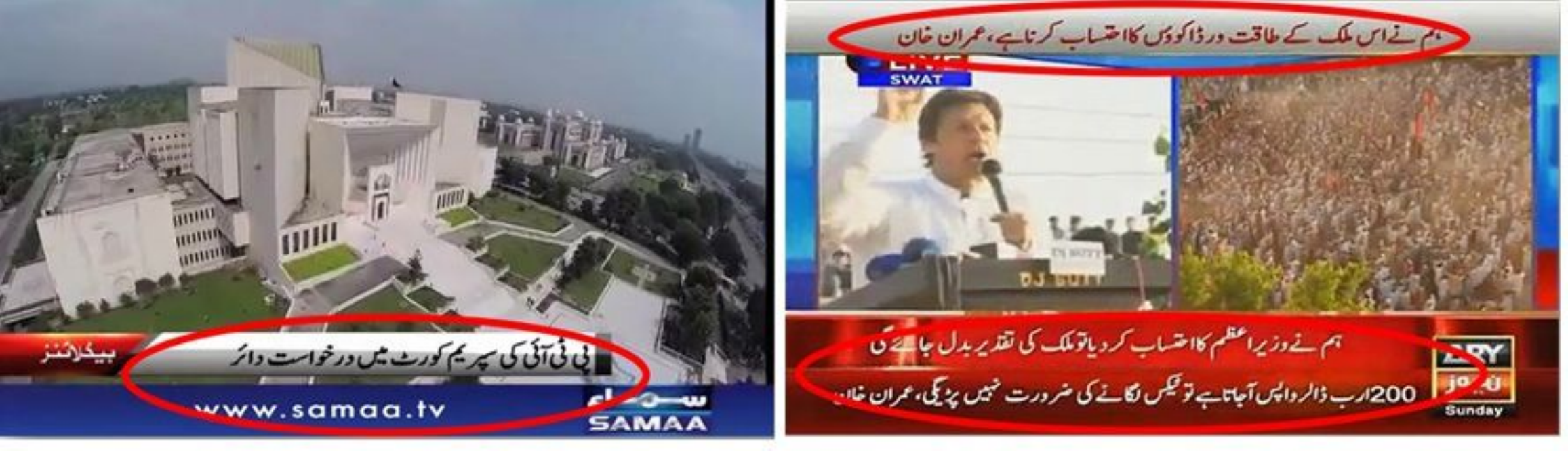}
	\caption{Examples of Caption Text}
		\label{fig:captext}
\end{figure}

The textual content in video can be categorized into two broad classes, scene text and caption text. Scene text (Figure~\ref{fig:scenetext}) is captured through camera during the video recording process and may not always be correlated with the content. Examples of scene text include advertisement banners, sign boards, text on a T-shirt etc. Scene text is commonly employed for applications like robot navigation and assistance systems for the visually impaired. Artificial or caption text (Figure~\ref{fig:captext}) is superimposed on video and typical examples include News tickers, movie credits, score cards, names of anchors etc. Caption text is generally correlated with the video content and is mostly applied for semantic retrieval of videos. \\

The key components of a textual content based indexing and retrieval system include detection of text regions~\cite{mirza2018urdu}, extraction of text (segmentation from background)~\cite{ali2019}, identification of script (for multi-script videos)~\cite{ali2016} and finally recognition of text (through a video OCR)~\cite{hayat2018ligature}. Among these, we focus on detection of text in the current study. Detection of text can be carried out using unsupervised~\cite{baran2018automated, dai2018scene, banerjee2013robust}, supervised~\cite{xu2018end, he2018single, liao2017textboxes} or hybrid~\cite{mirza2018urdu, shrivastava2017learning} approaches. Unsupervised text detection employs image analysis techniques to discriminate between text and non-text regions. Supervised methods, on the other hand, involve training a learning algorithm with examples of text and non-text regions to discriminate between the two. In some cases, a combination of the two techniques is employed where the candidate text regions identified by unsupervised methods are validated through a supervised approach. \\

This paper presents a comprehensive framework for video text detection in a multi-script environment. Though we primarily target cursive caption text, since video frames frequently contain text in more than one script, text in the Roman script is also detected by the proposed technique. The key highlights of this study are outlined in the following.

\begin{itemize}
    \item Development of a comprehensive dataset of video images with ground truth information supporting evaluation of detection and recognition tasks.
    \item Adaptation of various deep learning based object detectors for detection of textual content.
    \item Combination of text detection and script identification in a single end-to-end system.
    \item Validation of proposed technique through an extensive series of experiments and a comprehensive performance comparison of various detectors.
\end{itemize}

The paper is organized as follows. In the next section, we present an overview of the current state-of-the-art on detection of textual content in videos. In Section~\ref{sec:Dataset}, we introduce the dataset developed in our study along with the ground truth information. Section~\ref{sec:Methodology} presents the details of the proposed framework while Section~\ref{sec:Experiments} presents the experimental protocol, the realized results and the corresponding discussion. Finally, Section~\ref{sec:Conclusion} concludes the paper with a discussion on open challenges on this subject.

\section{Background}\label{sec:lit}
Detection of textual content in videos, images, documents and natural scenes has been investigated for more than four decades. The domain has matured progressively over the years starting with trivial image analysis based systems to complex end-to-end learning based systems. We discuss notable contributions to text detection in the following while detailed surveys on the problem (and related problems) can be found in~\cite{Ye2015,wang2016,yin2016text,zhang2013,sharma2012recent}. \\

Text detection refers to localization of textual content in images. Techniques proposed for detection of text are typically categorized into unsupervised and supervised approaches. While unsupervised approaches primarily rely on image analysis techniques to segment text from background, supervised methods involve training a learning algorithm to discriminate between text and non-text regions.\\

Unsupervised text detection techniques include edge-based methods~\cite{baran2018automated,dai2018scene,banerjee2013robust,huang2013scene,Jamil:2011} which (assume and) exploit the high contrast between text and its background; connected component based methods~\cite{Kiran:2012,koo2013scene,lee2010scene,pan2011hybrid} which mostly rely on the color/intensity of text pixels and texture-based methods~\cite{XiangBai2016,Huang:2011} which consider textual content in the image as a unique texture that distinguishes itself from the non-text regions. Texture based methods have remained a popular choice of researchers and features based on Gabor filters~\cite{gabor1946}, wavelets~\cite{ye2005fast}, curvelets~\cite{joutel2007curvelets}, local binary patterns (LBP)~\cite{anthimopoulos2010two}, discrete cosine transformation (DCT)~\cite{zhong2000automatic}, histograms of oriented gradients (HoG)~\cite{dalal2005histograms} and Fourier transformation~\cite{shivakumara2010new} have been investigated in the literature. Another common category of techniques includes color-based methods~\cite{yi2012localizing,Yi:2007,Shivakumara:2010} which are similar in many aspects to the component-based methods and employ color information of text pixels to distinguish it from non-text regions. \\

Supervised approaches for detection of textual content typically employ state-of-the-art learning algorithms which are trained on examples of text and non-text blocks either using pixel values or by first extracting relevant features. Classifiers like naive Bayes~\cite{shivakumara2012multioriented}, Support Vector Machine~\cite{Zhen:2009}, Artificial Neural Network~\cite{Yin:2013,mirza2018urdu} and Deep Neural Networks~\cite{lecun2015deep} have been investigated for this problem over the years. \\

In the recent years, deep learning based solutions have been widely applied to a variety of recognition problems and have outperformed the traditional techniques. Among deep learning based techniques adapted for text detection, Huang et al.~\cite{huang2014robust} employed sliding windows with CNNs to detect textual regions in low resolution scene images. Likewise, fully convolutional networks are explored for detection of textaul regions in~\cite{zhang2016multi} and the technique is evaluated on various ICDAR datasets. A similar work is presented by Gupta et al.~\cite{gupta2016synthetic} where CNNs are trained using synthetic data for detection of text at multiple scales from natural images. Another method called `SegLink', is proposed in~\cite{shi2017detecting} that relies on decomposing the text into segments (oriented boxes of words or lines) and links (connecting two adjacent segments). The segments and links are detected using fully convolutional networks at multiple scales and combined together to detect the complete text line. In~\cite{tian2016detecting} vertical anchor based method is reported that predicts text and non-text scores of fixed size regions and reports high detection performance on the ICDAR 2013 and ICDAR 2015 datasets. In another notable work, Wang et al.~\cite{wang2018crf} present a framework based on conditional random field (CRF) to detect text in scene images. Authors define a cost function by considering the color, stroke, shape and spatial features with CNN for effective detection of textual regions.\\

%A major development contributing to the current fame of deep learning was the application of Convolutional Neural Networks (CNNs) by Krizhevsky et al.~\cite{krizhevsky2012imagenet} on the ImageNet Large Scale Visual Recognition competition~\cite{russakovsky2015imagenet}, which greatly reduced the error rates. Since then, CNNs are considered to be state-of-the-art feature extractors and classifiers~\cite{simonyan2014very,szegedy2015going} and have been applied to a number of recognition tasks~\cite{bouchain2006character,uijlings2013selective,farfade2015multi}. While traditional CNNs are typically employed for object classification, Region-based Convolutional Networks (R-CNN)~\cite{girshick2014rich} and their further enhancements Fast R-CNN~\cite{girshick2015fast} and Faster R-CNN~\cite{ren2015faster} represent state-of-the-art object detectors. In addition to different variants of R-CNN, a number of new architectures have also been proposed in the recent years for real time object detection. The most notable of these include YOLO (You Only Look Once)~\cite{redmon2016you} and SSD MultiBox (Single Shot Detector)~\cite{liu2016ssd}. These object detectors can be fine-tuned with textual data to serve as text detectors and provide very good results.\\

Among other end-to-end trainable deep neural networks based systems, Liao et al.~\cite{liao2017textboxes} present a system called `TextBoxes' which detects text in natural images in a single forward pass network. The technique was later extended to `TextBoxes++' and was evaluated on four public databases outperforming the state-of-the-art. He et al.~\cite{he2018single} improved the convolutional layer of CNNs to detect text with arbitrary orientation. EAST~\cite{zhou2017east}, is another well-known scene text detector that provides promising results in challenging scenarios. In another study~\cite{xu2018end}, an ensemble of Convolutional Neural Networks (CNNs) is trained on synthetic data to detect video text in East Asian languages.\\

The literature is relatively limited once it comes to detection of cursive caption text. Among one of the preliminary works, Jamil et al.~\cite{jamil2011edge} exploit edge based features with morphological processing to detect Urdu caption text from a small set of 150 video frames. The same study was extended by Raza et al.~\cite{siddiqi2012database} and evaluated on a larger set of 1000 video frames reporting a recall of 0.80. The dataset of images termed as IPC Artificial Text dataset~\cite{siddiqi2012database} was also made publicly available. In a later study~\cite{raza2013multilingual} by the same group, the authors proposed a cascaded framework of spatial transforms to detect caption text in five different scripts including Arabic and Urdu. In a relatively recent work on detection of Arabic caption text, Zayene et al.~\cite{Oussama2016} employ a combination of stroke width transform with a convolutional auto-encoder and evaluate the technique on a publicly available dataset AcTiV-DB~\cite{Oussama2015}. In one of our previous works~\cite{mirza2018urdu}, we investigated a combination of image analysis techniques with textural features to detect textual regions in video frames and realized an F-measure of 0.80 on 1000 images.\\

Summarizing, it can be concluded that the problem of text detection has been dominated by the application of different deep learning based techniques in the recent years. The availability of benchmark datasets has also contributed to the rapid developments in this area. While detection of text in languages based on the Latin alphabet has received significant research attention and is very much mature, detection of cursive text still remains a relatively less addressed and challenging issue. Development of a (generic) text detector that could work in multi-script environments also remains an open problem.\\

In the next section, we introduce the dataset that has been collected and labeled as a part of this study.

\section{Dataset}\label{sec:Dataset}
Availability of labeled datasets is of utmost importance for algorithmic development and evaluation of any computerized system. From the perspective of Urdu caption text detection, a dataset of 1000 labeled video frames has been made publicly available~\cite{siddiqi2012database}. A collection of 1000 frames, however, seems to be very small to generalize the findings for practical applications. We, therefore, collected and labeled a comprehensive dataset of video frames allowing evaluation of text detection and text recognition tasks. We collected a set of 46 videos from four different News channels in Pakistan. All videos are recorded at a resolution of $900 \times 600$ and a frame rate of 25 fps. Frames in these video contain textual content in two languages, (cursive) Urdu and English. The collected video frames are labeled from two perspectives, detection and recognition. For detection, the bounding rectangle of all text regions in a frame is labeled and stored. Similarly, for recognition, the transcription of each text line is stored as ground truth.

In the literature, several evaluation metrics have been proposed to evaluate the performance of text detection systems~\cite{jamil2011edge,lucas2005icdar,wolf2006object}.  In our system, for evaluation of the text detection module, we employ the most commonly used area based precision and recall measures reported in~\cite{jamil2011edge} and defined as follows.\\

Let $A_E$ be the estimated text area given by the system and $A_T$ be the ground truth text area, then the precision $P$ and recall $R$ are defined as:
\begin{equation}
	P~ =~\frac{~A_E~ \cap~ A_T}{A_E}
\end{equation}

\begin{equation}
	R~ =~ \frac{~A_E~ \cap~ A_T}{A_T}
\end{equation}
The precision and recall measures can be combined in a single F-measure as follows.

\begin{equation}
	F~=~ \frac{~2~ \times ~Precision~ \times~ Recall~ }{~Precision ~+~ Recall}
\end{equation}

\begin{figure*}[!ht]
	\centering
		\includegraphics[width=0.95\textwidth]{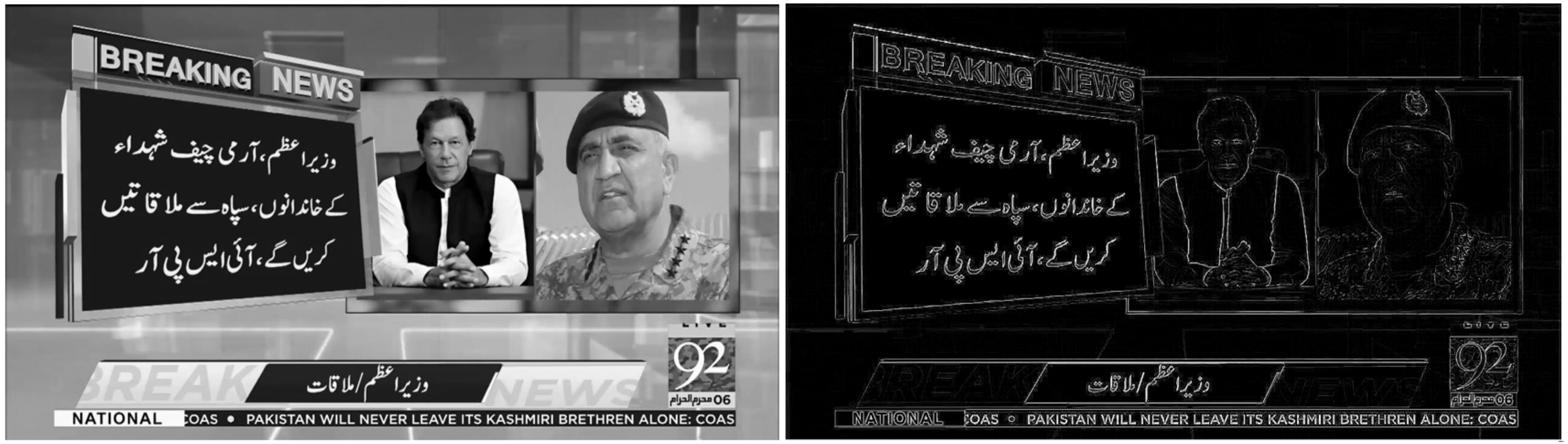}
	\caption{Text regions in an image and the corresponding ground truth image}
	\label{fig:textreg}
\end{figure*}

The same idea can be extended to multiple images by simply summing up area of intersection and dividing by the total ground truth area (in $N$ images) for recall and the total detected area for precision. To compute these measures, for each frame, we need to store the actual location of the textual content. The text detected automatically by the system can then be compared with the ground truth text regions to compute precision, recall and F-measure. The idea is illustrated in Figure~\ref{fig:textreg}. Figure~\ref{fig:textreg}-a illustrates an example where the text regions detected by the system are shown while Figure~\ref{fig:textreg}-b illustrates the ground truth text locations for the given frame. The detected and ground truth text regions can be compared to compute the metrics defined earlier and quantify the detection performance.\\

\begin{figure*}[!ht]
	\centering
		\includegraphics[width=0.95\textwidth]{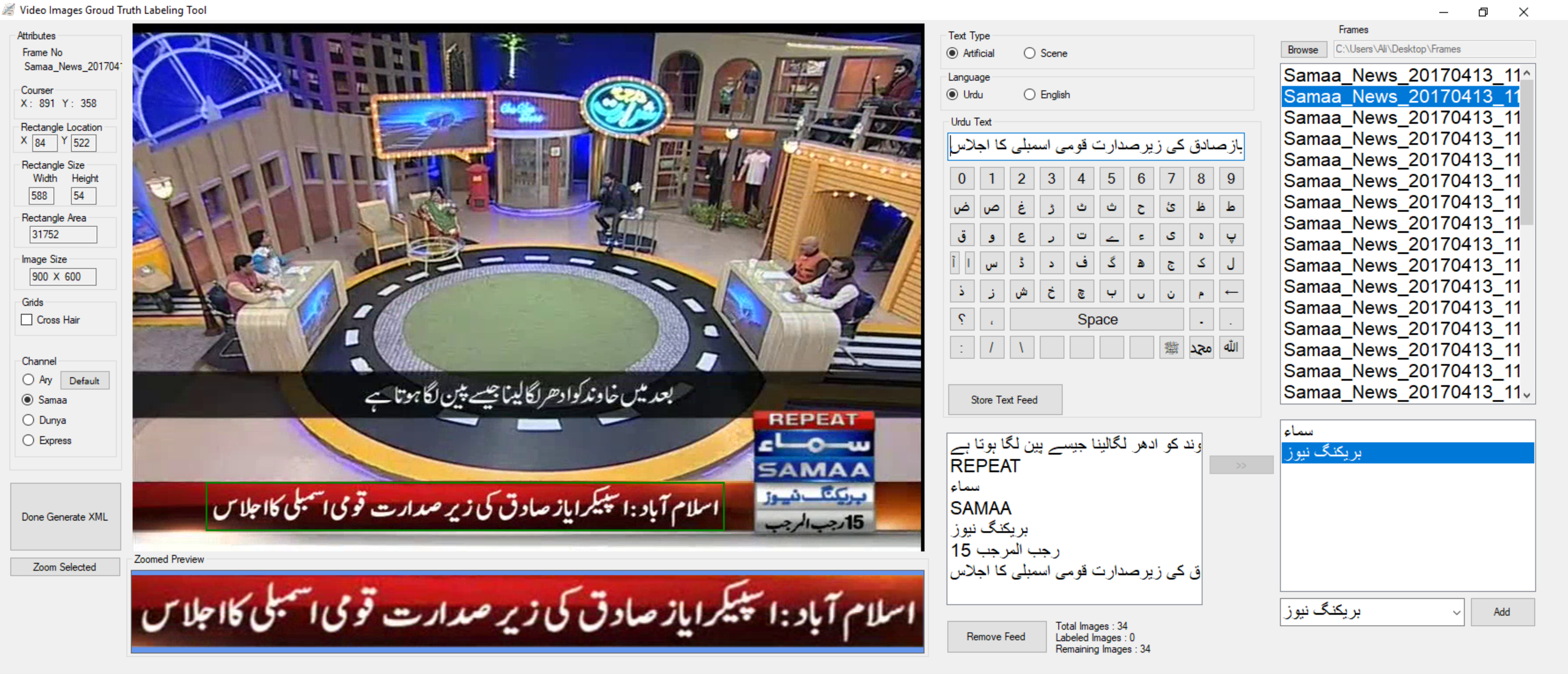}
		\caption{Screen shot of ground truth labeling tool for text data}
	\label{fig:GTTool}
\end{figure*}

\begin{figure*}[!ht]
	\centering
		\includegraphics[width=0.95\textwidth]{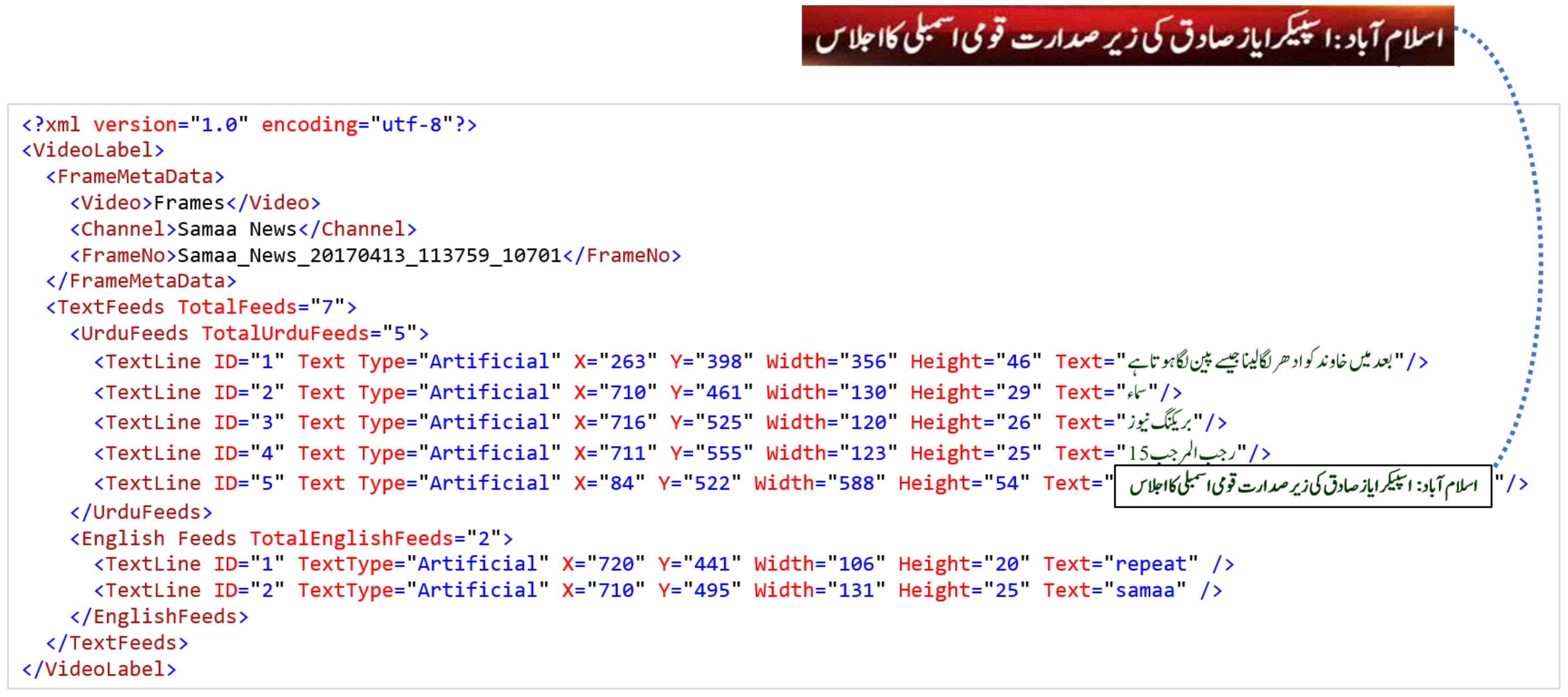}
	\caption{An XML file containing ground truth information of a frame}
		\label{fig:XMLScreen}
\end{figure*}

To facilitate the labeling process and standardize the ground truth data, a comprehensive labeling tool has been developed that allows storing the location of each textual region in a frame along with its ground truth transcription. The location is stored in terms of the $x$ and $y$ coordinates of the top left of the bounding box along its $width$ and $height$. The ground truth information of each frame is stored as an XML file that comprises frame meta data and the information on textual regions. A screen shot of the labeling tool is presented in Figure~\ref{fig:GTTool} while the ground truth information of an example frame is illustrated in Figure~\ref{fig:XMLScreen}.\\

It is known that videos typically contain 25--30 frames per second; consequently, successive frames in a video contain redundant information (both visual and textual content).  From the view point of automatic analysis systems, frames with unique content are of interest. Hence, each single video frame does not need to be labeled as major proportions of such frames will have exactly the same textual information. In our study, we have extracted more than 11,000 frames from videos with an attempt to have as much unique text as possible. The statistics of videos, frames and text lines of our dataset are presented in Table~\ref{tab:channelStats}. 
Inspired from the Arabic caption text dataset AcTiV-DB~\cite{Oussama2015}, we have named the our dataset UTiV (Urdu Text in Video). The dataset along with its ground truth has also been made publicly available\footnote{http://cbvir.media-tics.net/} to support quantitative evaluation of text detection and recognition tasks.\\

\begin{table*}[!ht]
\centering
\caption{Statistics of labeled video frames}
\label{tab:channelStats}
%\footnotesize
\begin{tabular}{llcccc}
\hline 
\textbf{S\#} & \textbf{Channel} & \textbf{Videos} & \textbf{Labeled Images} & \textbf{Urdu Lines} & \textbf{English Lines}\\
\hline
1 & Ary News & 7 & 3,206 & 10,250 & 3,605\\
2 & Samaa News & 13 & 2,503 & 10,961 & 4,411\\
3 & Dunya News & 16 & 3,059 & 10,723 & 8,861\\
4 & Express News & 10 & 2,424 & 8,536 & 6,755\\
& \textbf{Total} & \textbf{46}& \textbf{11,203} &  \textbf{40,470} & \textbf{23,632}\\
\hline 
\end{tabular}
\end{table*}

\section{Methods}\label{sec:Methodology}
This section presents the details of detecting textual content from video frames. Detection relies on adapting object detectors based on deep convolutional neural networks for text regions. Once the text is detected, script of the detected text is identified by employing the ConvNets in a classification framework. Subsequently, text detection and script identification are combined in a single end-to-end system that detects the textual content along with its script. Details are presented in the following sections.

\subsection{Deep Learning based Object Detectors}
Deep neural networks enjoy a renewed interest of the machine learning community thanks primarily to the availability of high performance computing hardware (GPUs) as well as large data sets to train these systems. A major development contributing to the current fame of deep learning was the application of ConvNets by Krizhevsky et al.~\cite{krizhevsky2012imagenet} on the ImageNet Large Scale Visual Recognition competition~\cite{russakovsky2015imagenet}, which greatly reduced the error rates. Since then, CNNs are considered to be state-of-the-art feature extractors and classifiers~\cite{simonyan2014very, szegedy2015going} and have been applied to a variety of recognition tasks~\cite{bouchain2006character, uijlings2013selective, farfade2015multi}. \\

%In addition to classification, CNNs have also been adapted for object detection and are known to outperform the conventional computer vision algorithms for detecting and localizing objects in images. Inspired by their robustness, we have chosen to adapt deep learning based object detectors for detection of textual content in the video frames. We first present an overview of the well-known object detectors followed by details on how they are adapted for detection of textual content in video frames. \\

%
While traditional CNNs are typically employed for object classification, Region-based Convolutional Networks (R-CNN)~\cite{girshick2014rich} and their further enhancements Fast R-CNN~\cite{girshick2015fast} and Faster R-CNN~\cite{ren2015faster} adapt CNNs for object detection. In addition to different variants of R-CNN, a number of new architectures have also been proposed in the recent years for real time object detection. The most notable of these include YOLO (You Only Look Once)~\cite{redmon2016you} and SSD (Single Shot Detector)~\cite{liu2016ssd}. Each of these object detectors can be trained to detect $C$ object classes (plus one for the background). The output of the detector is the location of the bounding box (four coordinates) containing one of the $C$ classes as well as the class confidence score.\\

In our study, for detection of textual content in a given frame, we investigated a number of CNN based object detectors. Although, many object detectors are trained with thousands of class examples and provide high accuracy in detection and recognition of different objects, these object detectors can not be directly applied to identify text regions in images. These models have to be tuned to the specific problem of discrimination of text from non-text regions. The convolutional base of these models can be trained from scratch or, known pre-trained models can be fine-tuned by training them on text and non-text regions. In our study, we investigated the following object detectors for localization of text regions.

\begin{itemize}
	\item Faster R-CNN
	\item Region-Based Fully Convolutional Networks (R-FCN) 
	\item Single Shot Detector (SSD) 
	\item You Only Look Once (YOLO)
\end{itemize}

For completeness, we provide a brief overview of these object detectors in the following sections.

\subsubsection{Faster R-CNN}
\begin{figure*}[!ht]
	\centering
		\includegraphics[width=0.95\textwidth]{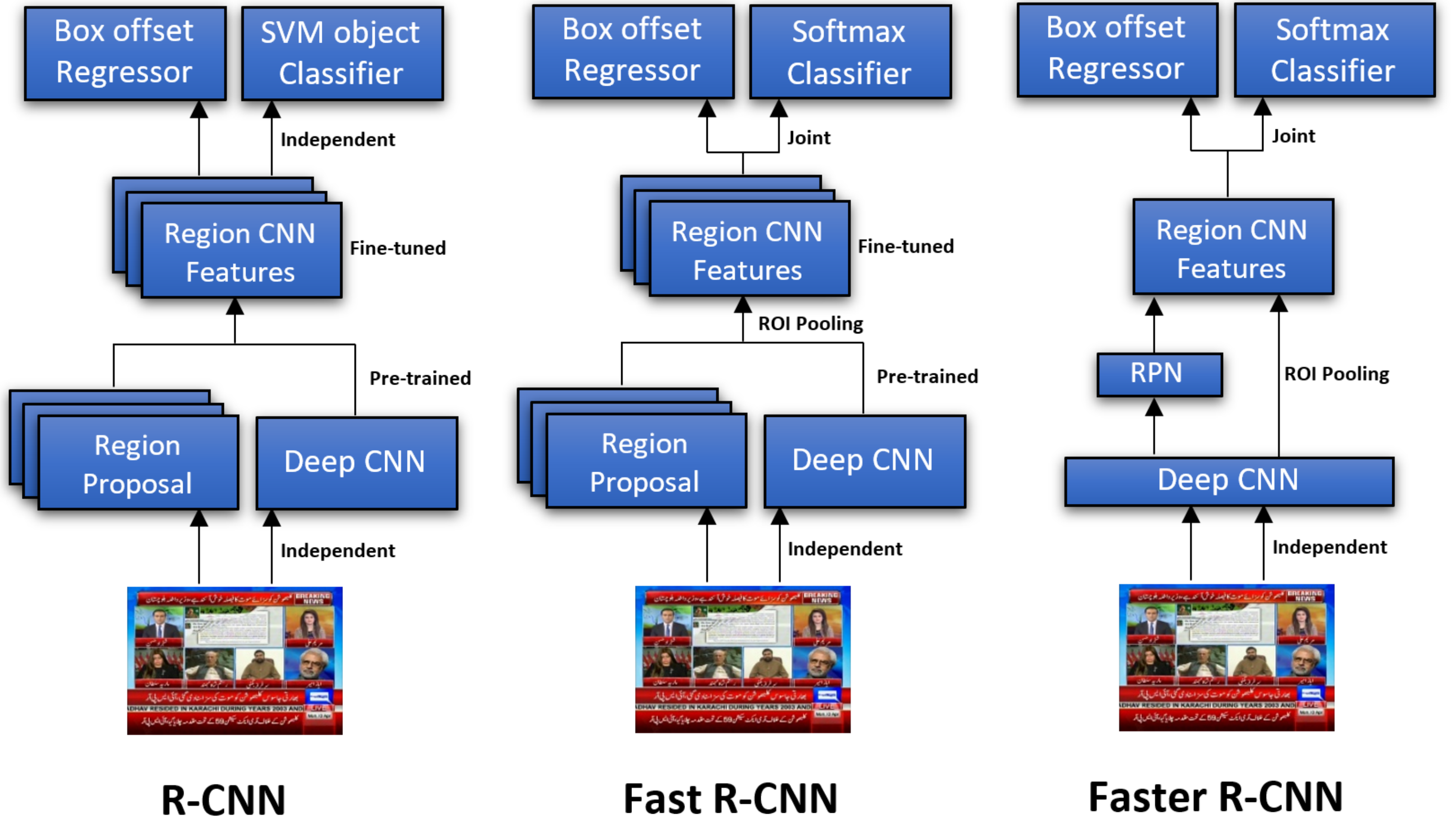}
	\caption{Summary of R-CNN Family based Object Detectors}
\label{fig:frcnn}
	
\end{figure*}

Faster R-CNN~\cite{ren2015faster} is an enhanced version of its predecessors R-CNN~\cite{girshick2014rich} and Fast R-CNN~\cite{girshick2015fast}. Each of these detectors exploits the powerful features of ConvNets for object localization as well as classification. R-CNN was one of the first attempts to apply ConvNets for object detection. An R-CNN scans the input image for potential objects using Selective Search~\cite{uijlings2013selective} that generates around 2,000 region proposals. Each of these region proposals is then fed to a CNN for feature extraction. The output of the CNN is finally employed by an SVM to classify the object and a linear regressor to tighten the bounding box. R-CNN was enhanced in terms of training efficiency by extending it to Fast R-CNN~\cite{girshick2015fast}. In Fast R-CNN, rather than separately feeding each region proposal to the ConvNet, convolution is performed only once on the complete image and the region proposals are projected on the feature maps. Furthermore, the SVM in R-CNN was replaced by a softmax layer extending the network to predict the class labels rather than using a separate model. While Fast R-CNN significantly reduced the time complexity of the basic R-CNN, a major bottleneck was the selective search algorithm to generate the region proposals. This was addressed through Region Proposal Network (RPN) in Faster R-CNN~\cite{ren2015faster} which shares convolutional features with the detection network. RPN predicts region proposals which are then fed to the detection network to identify the object class and refine the bounding boxes produced by the RPN. A summary of various R-CNN models in presented in Figure~\ref{fig:frcnn}.

\subsubsection{You Only Look Once (YOLO)}
YOLO~\cite{redmon2016you} takes a different approach to object detection primarily focusing on improving the detection speed (rather than accuracy). As the name suggests, YOLO employs a single pass of the convolutional network for localization and classification of objects from the the input images. The input image is divided into a grid and an object is expected to be detected by the grid which holds the center of the object. Each cell in the grid predicts up to two bounding boxes (and class probabilities). The network comprises 24 convolutional and  fully connected layers. YOLO works in real time but in terms of accuracy, it is known to make significant localization errors in comparison to region based object detectors (Faster R-CNN for instance). 
YOLO was later enhanced to YOLO9000~\cite{redmon2017yolo9000} by introducing batch normalization, increasing the resolution of the input image (by a factor of 2) and introducing the concept of anchor boxes. YOLO9000 employs Darknet 19 architecture with 19 convolutional layers, 5 max pooling layers and a softmax layer for classification objects. Incremental improvements in YOLO v2 resulted in YOLO v3~\cite{redmon2018yolov3} that uses logistic regerssion to predict the score of objectness for each bounding box. Furthermore, it employs class-wise logistic classifiers (rather than softmax) allowing multi-label classification.

\subsubsection{Single Shot Detector (SSD)}
Unlike the R-CNN series object detectors which require two shots to detect objects in an image, Single Shot Multi-box Detector~\cite{liu2016ssd}, as the name suggests, requires a single shot to detect objects (similar to YOLO). SSD relies on the idea of default boxes and multi-scale predictions and directly applies bounding box regression to the default boxes without generating the region proposals. Detection at multiple scales are handled by exploiting the feature maps of different convolutional layers corresponding to different receptive fields in the input image. The architecture has an input size of $300 \times 300 \times 3$ and primarily builds on the VGG-16 architecture discarding the fully connected layers. VGG-16 is used as base network mainly due to its robust performance of image classification tasks. The bounding box regression technique of SSD is inspired by~\cite{szegedy2015going} while the MultiBox relies on priors, the pre-computed fixed size bounding boxes. The priors are selected in such a way that their Intersection over Union ratio (with ground truth objects) is greater than 0.5. The MultiBox starts with the priors as predictions and attempt to regress closer to the ground truth bounding boxes. SSD works in real time but requires images of fixed square size and is known to miss small objects in the image.

\subsubsection{Region-Based Fully Convolutional Networks (R-FCN) }
R-FCN~\cite{dai2016r} builds on the idea of increasing the detection accuracy by maximizing the shared calculations. R-FCN generates position-sensitive score maps to represent different relative positions of an object. An object is represented by $k^2$ relative positions dividing it into a grid of size $k \times k$. A ConvNet (ResNet in the original R-FCN paper) sweeps the input image and an additional fully convoltional layer produces the position-sensitive scores in $k^2 \times (C+1)$ score maps where $C$ is the number of classes plus 1 class for the background. A fully convolutional proposal network generates regions of interest which are divided in $k^2$ bins and the corresponding class probabilities are obtained from the score maps. The scores are averaged to convert the $k^2 \times (C+1)$ values into a one dimensional $(C+1)$ sized vector which is finally fed to the softmax layer for classification. Localization is carried out using the bounding box regression similar to other object detectors. R-FCN speeds up the detection in comparison to Faster R-CNN but compared to other Single Shot methods, it requires more computational resources.

\subsection{Adapting Object Detectors for Text Detection}
\begin{figure*}[!ht]
	\centering
		\includegraphics[width=1.0\textwidth]{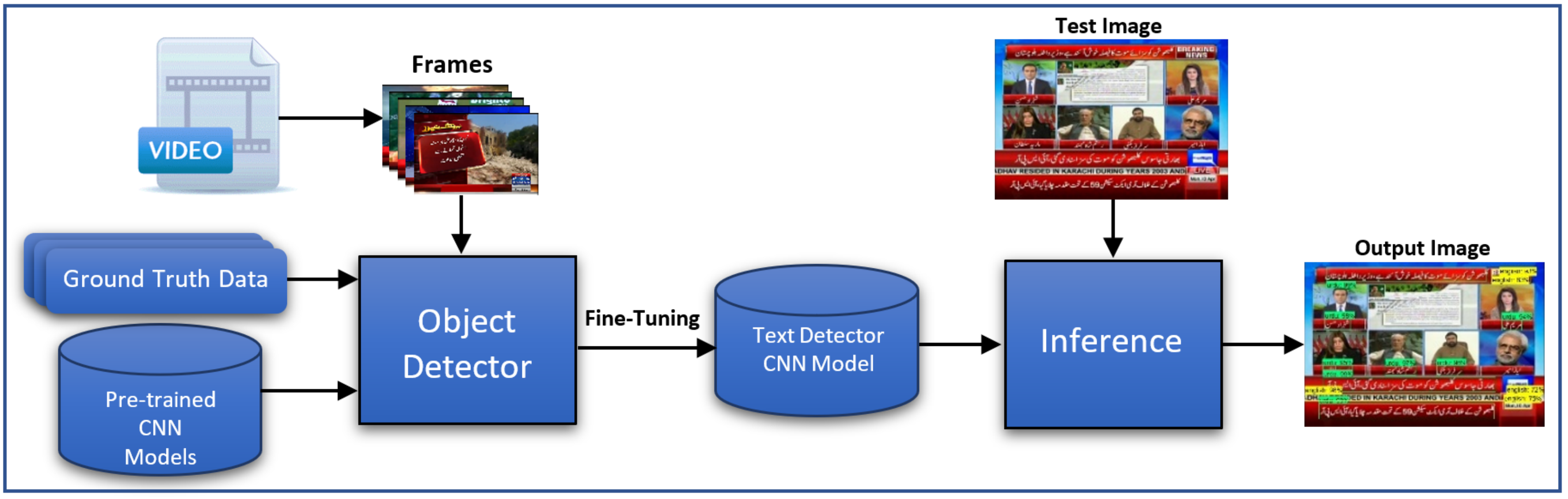}
	\caption{Overview of adapting object detectors for text detection}
\label{fig:traintestmethod}
\end{figure*}

In the context of object detection, the problem of text detection can be formulated as a two class problem. The text regions represent object of interest while the non-text regions need to be ignored. The object detectors discussed in the previous sections are adapted for text detection using two pre-trained models, ResNet 101~\cite{he2016deep} and Inception v2~\cite{ioffe2015batch}. These models are trained on the large scale Microsoft COCO (Common Objects in Context) database~\cite{lin2014microsoft}. The database contains images of 91 different object types with a total of 2.5 million labeled instances in 328K images. The pre-trained network serves as starting point rather than random weight initialization and the network is made to learn the specific class labels (text or non-text) by continuing back propagation. The ground truth localization information of the textual regions in the video frames is employed for training the models, the overall workflow being illustrated in Figure~\ref{fig:traintestmethod}. \\

\begin{figure}[!ht]
	\centering
		\includegraphics[width=0.75\textwidth]{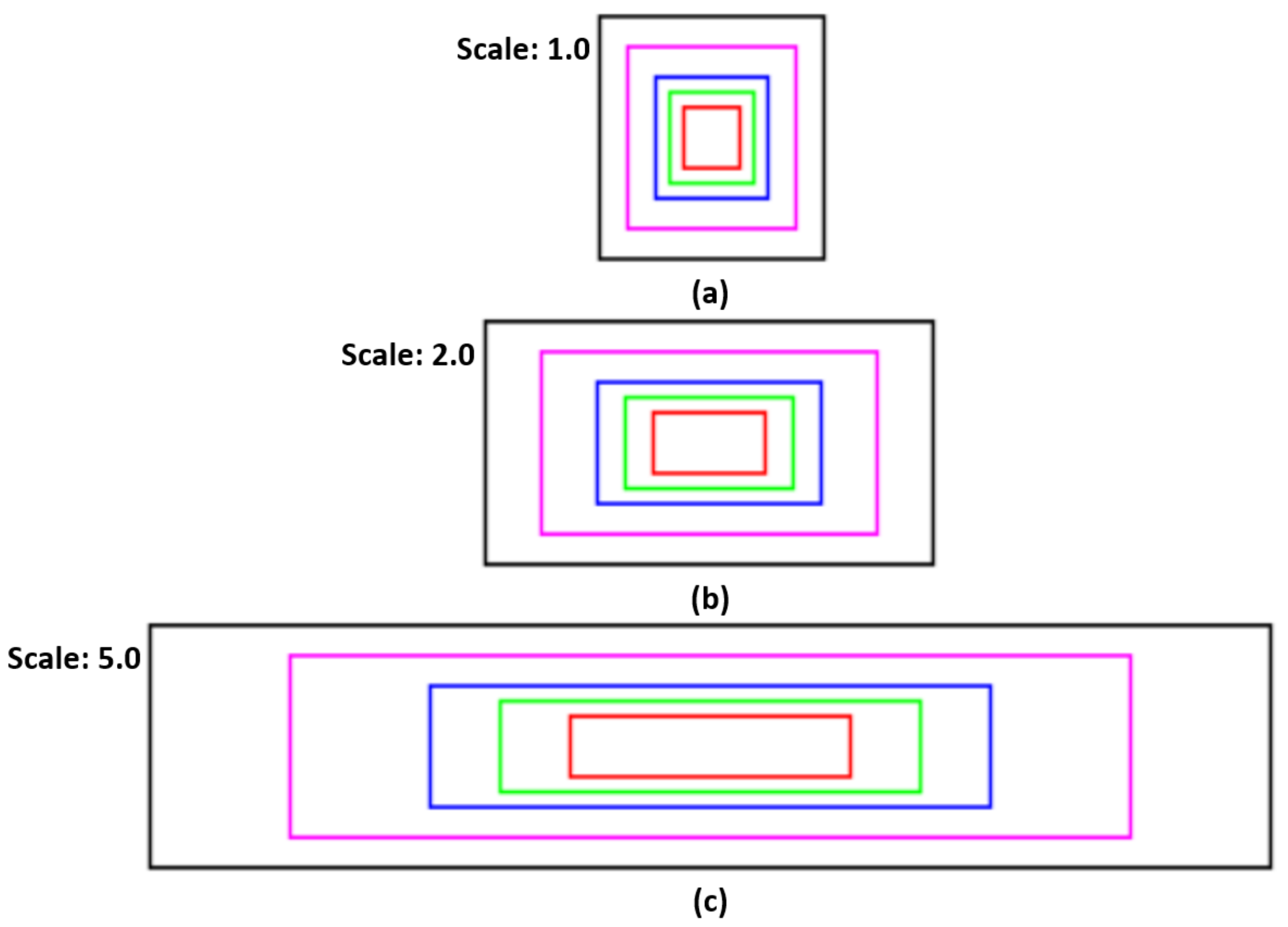}
	\caption{Anchor boxes (base size $256 \times 256$) at three scales ($1.0, 2.0, 5.0$) and five aspect ratios ($0.125, 0.1875, 0.25, 0.375, 0.50$)}
\label{fig:Scales}
\end{figure}

A critical aspect in employing object detectors for text detection is the choice of anchor boxes. The anchor boxes in all the detectors have been designed to detect general object categories. Text appearing in videos has specific geometric properties in terms of size and aspect ratio hence the default anchor boxes of the detectors need to be adapted to detect textual regions. We carried out a comprehensive analysis of the textual regions in terms of width, height and aspect ratios of the bounding boxes. As a result of this analysis we have chosen a base anchor of size $256 \times 256$. To each anchor box we apply three scales ($1.0, 2.0, 5.0$) and five aspect ratios ($0.125, 0.1875, 0.25, 0.375, 0.50$) as illustrated in Figure~\ref{fig:Scales}. Models are fine-tuned using the proposed anchor boxes and the effectiveness of these anchor boxes is validated through experimental study as presented in Section~\ref{sec:Experiments} of the paper.   \\

% scales : 5.0 2.0 1.0
% aspect ratios : 0.125, 0.1875, 0.25, 0.375, 0.5

\subsection{Script Identification}
\begin{figure*}[!ht]
	\centering
		\includegraphics[width=0.95\textwidth]{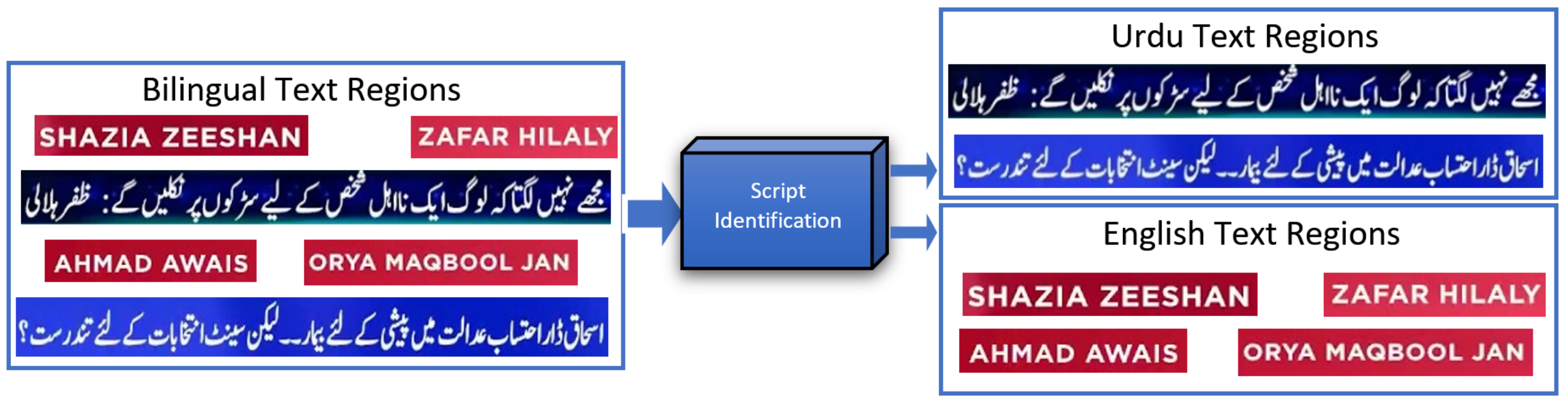}
	\caption{Script identification of detected text lines}
\label{fig:scriptIden}
\end{figure*}

As discussed earlier, we primarily target detection of cursive caption text. However, like many practical scenarios, video frames in our case contain bilingual textual content (Urdu \& English). Consequently, once the text is detected, we need to identify the script of each detected region (Figure~\ref{fig:scriptIden}) so that the subsequent processing of each type of script can be carried out by the respective recognition engine. For script identification, we employ CNNs in a classification framework (rather than detection). Urdu and English text lines are employed to fine-tune CNNs to discriminate between the two classes. Once trained, the model is able to separate text lines as a function of the script. Similar to detection, rather than training the networks from scratch, we fine-tune known pre-trained models (Inception and ResNet) to solve the two-class classification problem.

\subsection{Hybrid Text Detector \& Script Identifier}
Detection of text and identification of script, as discussed previously, can be implemented in a cascaded framework where the output of text detector is fed to the script identifier. A deep learning framework can be tuned to discriminate between text and non-text regions and the extracted text regions can be fed to a separate script recognition model that identifies the script of the detected text. This, however, introduces a bottleneck of training two separate networks. Furthermore, the cascaded solution also implies that errors in detection are propagated to the next step as well. We, therefore, propose to combine the text detector and script identifier in a single hybrid model. Rather than treating detection as a two-class problem (text and non-text), we consider it as a three class problem, i.e. non-text regions, English text and Urdu text. This not only avoids training two separate models but also eliminates the accumulation of errors in a cascaded solution. The superiority of the combined text detector and script identifier is also supported through quantitative evaluations as discussed in the next section.\\

\begin{figure}[!ht]
	\centering
		\includegraphics[width=0.75\textwidth]{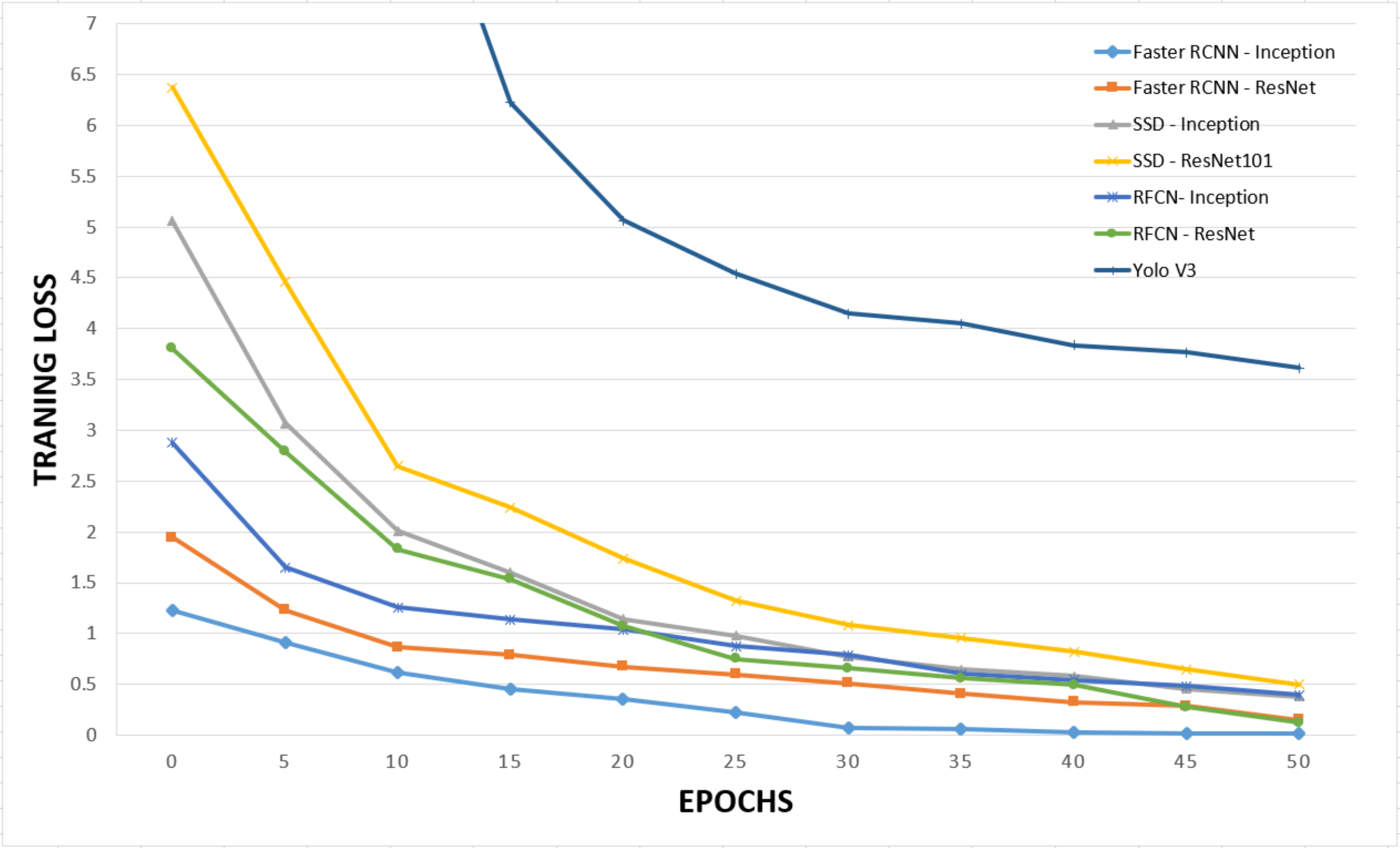}
	\caption{Training loss of various detectors -- Hybrid text detector and script identifier}
\label{fig:alllosses}
\end{figure}

All detectors are trained in an end-to-end manner with a multi-task objective function that combines the classification and regression losses. The evolution of training loss for the investigated detectors (with Inception and ResNet) is illustrated in Figure~\ref{fig:alllosses} where it can be seen that the loss begins to stabilize from 30 epochs on wards. 

\section{Experiments and Results}\label{sec:Experiments}
The detection performance is evaluated through a series of experiments carried out on the collected set of video frames. We first present the experimental protocol followed by the detection results of various object detectors. We then present the script identification results and the performance of the combined text detector and script identifier. Furthermore, performance sensitivity of the system as well as a comparison with state-of-the-art is also presented. 

\subsection{Experimental Settings}
As introduced in Section~\ref{sec:Dataset}, we collected a total of 11,203 video frames from four different News channel videos. The localization information of text regions in these frames is used to train and subsequently evaluate the text detection and script identification performance. The distribution of frames into training and test sets along with the number of text lines in each set is summarized in Table~\ref{tab:dataDist} while the details of detection performance are presented in the next section.

\begin{table}[!ht]
\centering
\caption{Data distribution for text detection experiments}
\label{tab:dataDist}
%\scriptsize

% \usepackage{multirow}
%\footnotesize
%\scriptsize
\begin{tabular}{lllll}
\hline %\cline{2-5}
                                       & \multicolumn{2}{c}{\textbf{Train}}                               & \multicolumn{2}{c}{\textbf{Test}}                                        \\ \hline%\cline{2-5} 
                                       & \multicolumn{1}{c}{\textbf{Frames}} & \multicolumn{1}{c}{\textbf{Lines}} & \multicolumn{1}{c}{\textbf{Frames}} & \multicolumn{1}{c}{\textbf{Line}} \\ \hline
\multicolumn{1}{l}{\textbf{Urdu}}    & \multirow{2}{*}{8,500}               & 31,321                              & \multirow{2}{*}{2,703}               & 9,149                              \\ %\cline{1-1} \cline{3-3} \cline{5-5} 
\multicolumn{1}{l}{\textbf{English}} &                                      & 16,207                              &                                      & 7,425                              \\ \hline
\multicolumn{1}{l}{\textbf{Total}}   &                                      & 49,046                     &                                      & 11,056                    \\ \hline

\end{tabular}
\end{table}

\subsection{Text Detection Results}
\begin{figure}[!ht]
	\centering
		\includegraphics[width=0.75\textwidth]{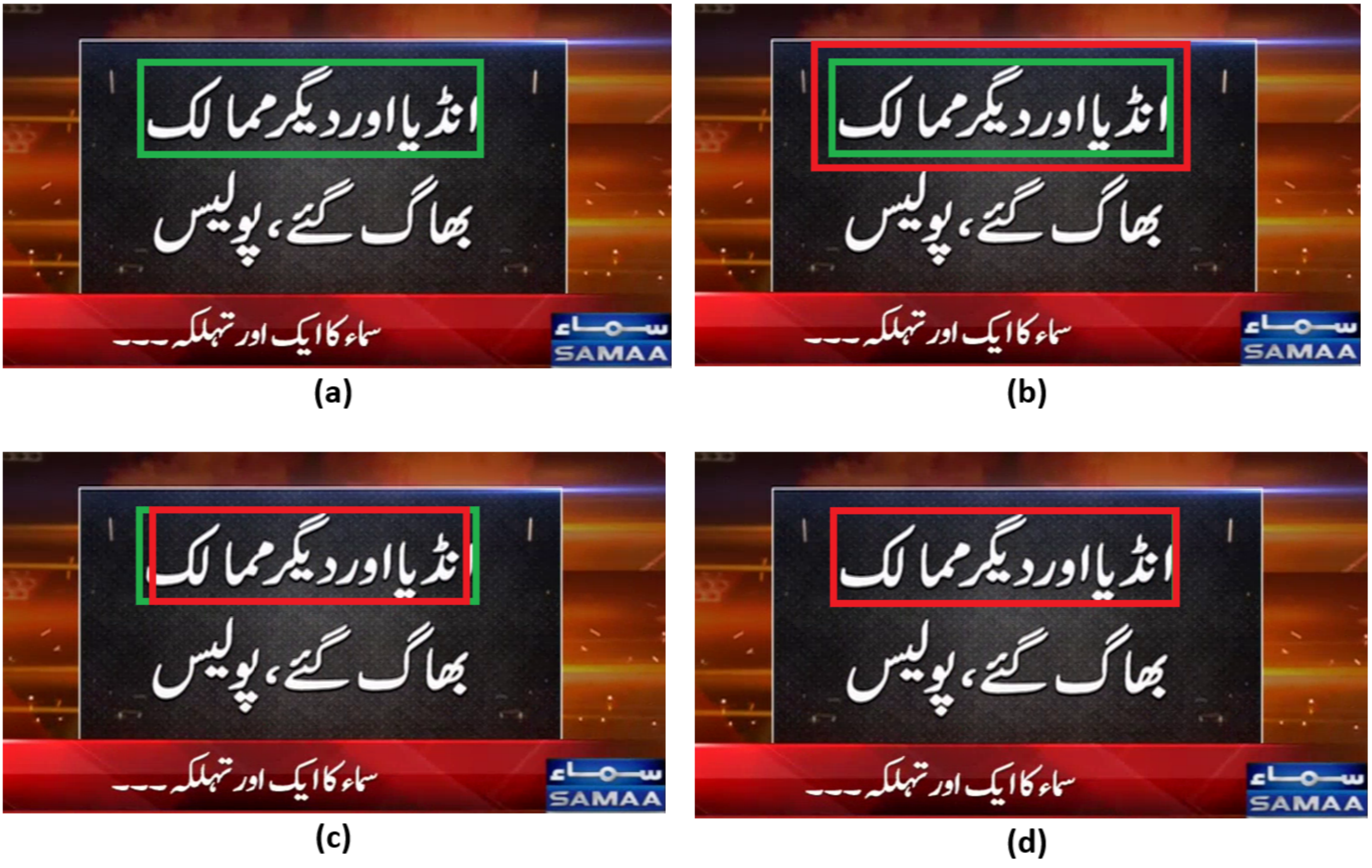}
	\caption{Computation of precision and recall (a):Ground Truth Bounding Box (b): Detected region is larger than ground truth (c):Detected region is smaller than ground truth (d):Detected region overlaps perfectly with the ground truth}
\label{fig:prerecall}
\end{figure}

\begin{figure}[!ht]
	\centering
		\includegraphics[width=0.85\textwidth]{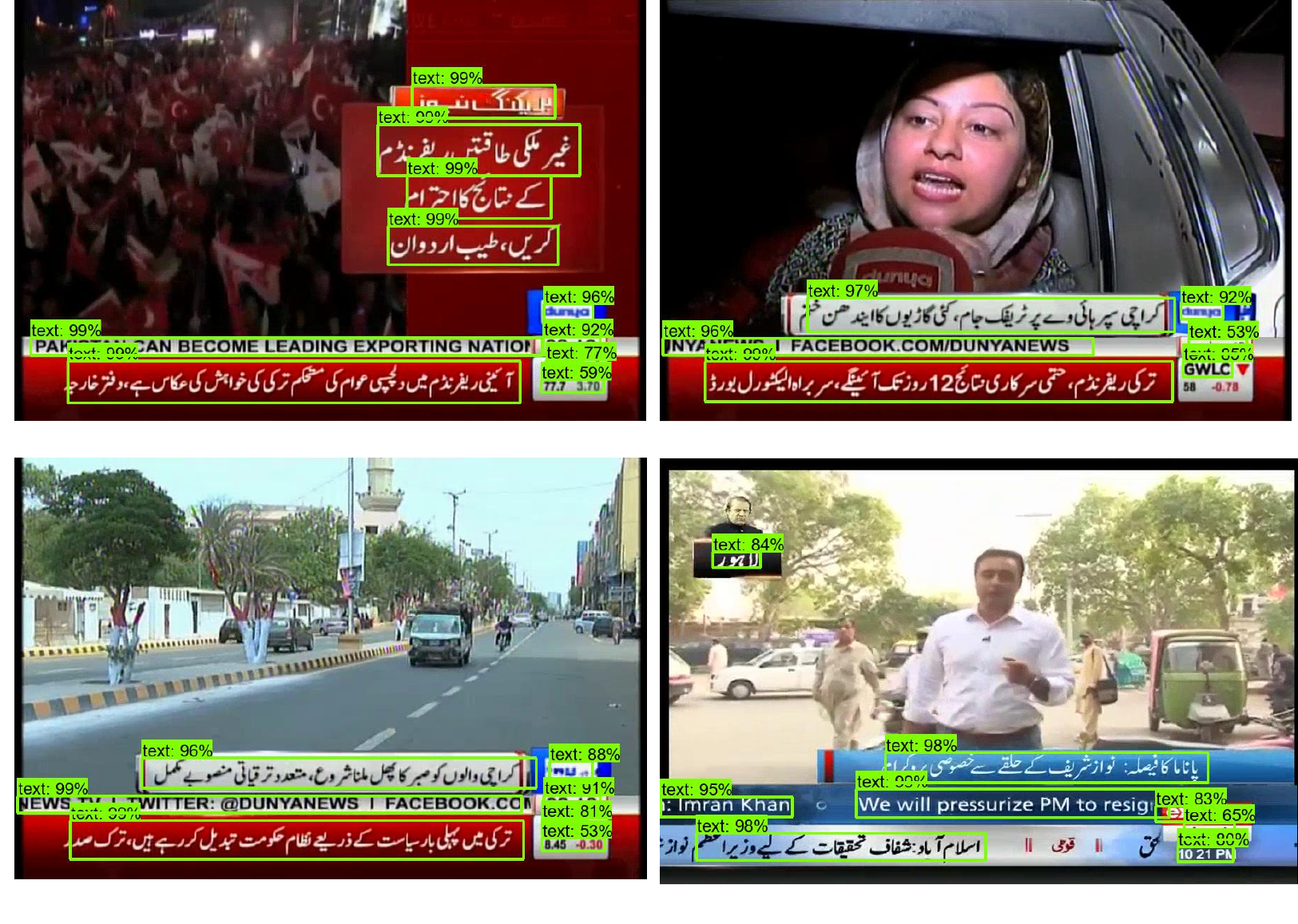}
	\caption{Text detection results on sample images (Faster R-CNN with Inception) }
\label{fig:detectionResultsSamp}
\end{figure}

Object detectors including Faster R-CNN, YOLO, SSD and R-FCN are adapted to detect textual content by fine-tuning the Inception and ResNet pre-trained models and changing the anchor boxes as discussed previously. Performance of each of these detectors in terms of precision, recall and F-measure is summarized in Table~\ref{tab:detectResText}. It can be seen that in all cases, detectors pre-trained on Inception outperform those trained on ResNet. Among various detectors, Faster R-CNN reports the highest F-measure of 0.90. The lowest performance is reported by Yolo reading an F-measure of 0.66. A comprehensive study on the trade-off between speed and accuracy of various object detectors is presented in~\cite{huang2017speed} and our findings on detection of text are consistent with those of~\cite{huang2017speed}. It is also important to recap that precision and recall are computed using area based metrics. As a result, if the detected bounding box is larger (smaller) than the ground truth, it results in penalizing the precision (recall) of the detector as illustrated in Figure~\ref{fig:prerecall}. The output of the Faster R-CNN based text detector for few sample frames in our dataset is illustrated in Figure~\ref{fig:detectionResultsSamp}. 
\\

\begin{table*}[!ht]
\centering
\caption{Text Detection Results}
\label{tab:detectResText}
%\scriptsize
%\footnotesize
\begin{tabular}{lcccccc}
\hline%\cline{2-7}
                                            & \multicolumn{3}{c}{\textbf{RestNet}}                     & \multicolumn{3}{c}{\textbf{Inception}}                   \\ \hline
\multicolumn{1}{l}{\textbf{Model}}        & \textbf{Precision} & \textbf{Recall} & \textbf{F-Measure} & \textbf{Precision} & \textbf{Recall} & \textbf{F-Measure} \\ \hline
\multicolumn{1}{l}{\textbf{SSD}}          & 0.83               & 0.71            & 0.77               & 0.82               & 0.77            & 0.80               \\ \hline
\multicolumn{1}{l}{\textbf{R-FCN}}        & 0.79               & 0.86            & 0.82               & 0.84               & 0.89            & 0.86               \\ \hline
\multicolumn{1}{l}{\textbf{Faster R-CNN}} & 0.82               & 0.90            & 0.85               & \textbf{0.86}      & \textbf{0.95}   & \textbf{0.90}      \\ \hline
\multicolumn{1}{l}{\textbf{Yolo}}         & -                  & -               & -                  & 0.63               & 0.69            & 0.66               \\ \hline
\end{tabular}
\end{table*}

\begin{figure}[!ht]
	\centering
		\includegraphics[width=0.85\textwidth]{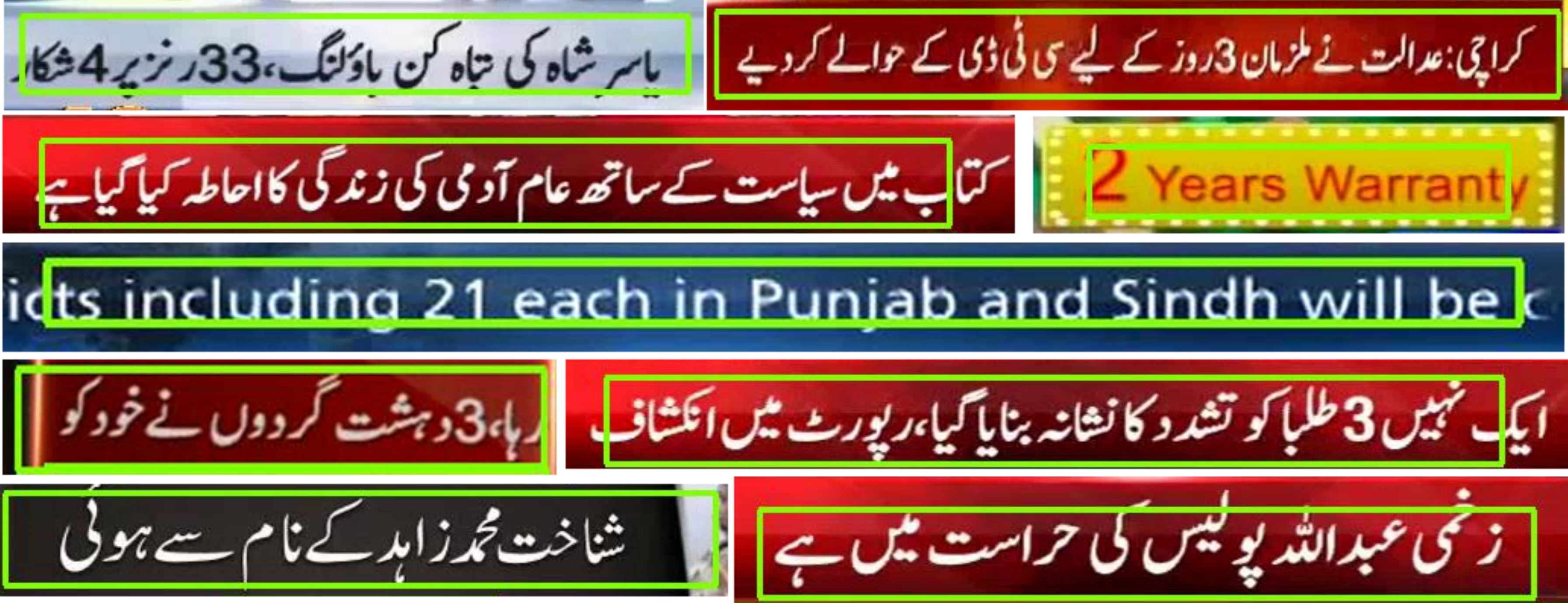}
	\caption{Imperfect Localization of Text Regions}
\label{fig:mistakes1}
\end{figure}

In an attempt to provide an insight into the detection errors, few of the errors are illustrated in Figure~\ref{fig:mistakes1}. It can be seen that in most cases, the detector is able to detect the textual region but the localization is not perfect i.e. in some cases the bounding box is larger (shorter) than the actual content leading to a reduced precision (recall). \\

\subsection{Script Identification Results}
For script identification, we employ the same distribution of frames into training and test sets as that of the detection protocol. Text lines from the video frames in the training set are employed to fine-tune the pre-trained ConvNets while the identification rates are computed on text lines from the frames in the test set. A total of $31,321$ Urdu and $16,207$ English text lines are used in the training set while the test set comprises $9,9149$ and $7,425$ text lines in Urdu and English respectively. The resulting confusion matrix is presented in Table~\ref{tab:siConMat} while the precision, recall and F-measure are summarized in Table~\ref{tab:sifmeasure}. It can be seen that the model was able to correctly identify the scripts with an accuracy of more than 94\%.\\

\begin{table}[!ht]

\centering

\caption{Script identification confusion matrix}
\label{tab:siConMat}
%\scriptsize
%\footnotesize
\begin{tabular}{lcc}
\hline
& \textbf{Urdu} & \textbf{English} \\
\hline%\cline{2-3}
\textbf{Urdu}  & 8763 & 386 \\
\textbf{English} & 551 & 6874 \\
\hline
\end{tabular}
\end{table}

\begin{table}[!ht]
\centering
\caption{Performance of Script Identification }
\label{tab:sifmeasure}
%\scriptsize
%\footnotesize
\begin{tabular}{lccc}
\hline

 &	\textbf{Precision} &	\textbf{Recall} & 	\textbf{F--Measure} \\ \hline

\textbf{Urdu} 	& 0.940 &	0.957 & 0.95\\
\textbf{English} & 0.946 & 0.925 & 	0.94 \\
\hline
\end{tabular}
\end{table}

\subsection{Hybrid Text Detection \& Script Identification Results}
\begin{figure*}[!ht]
	\centering
		\includegraphics[width=0.85\textwidth]{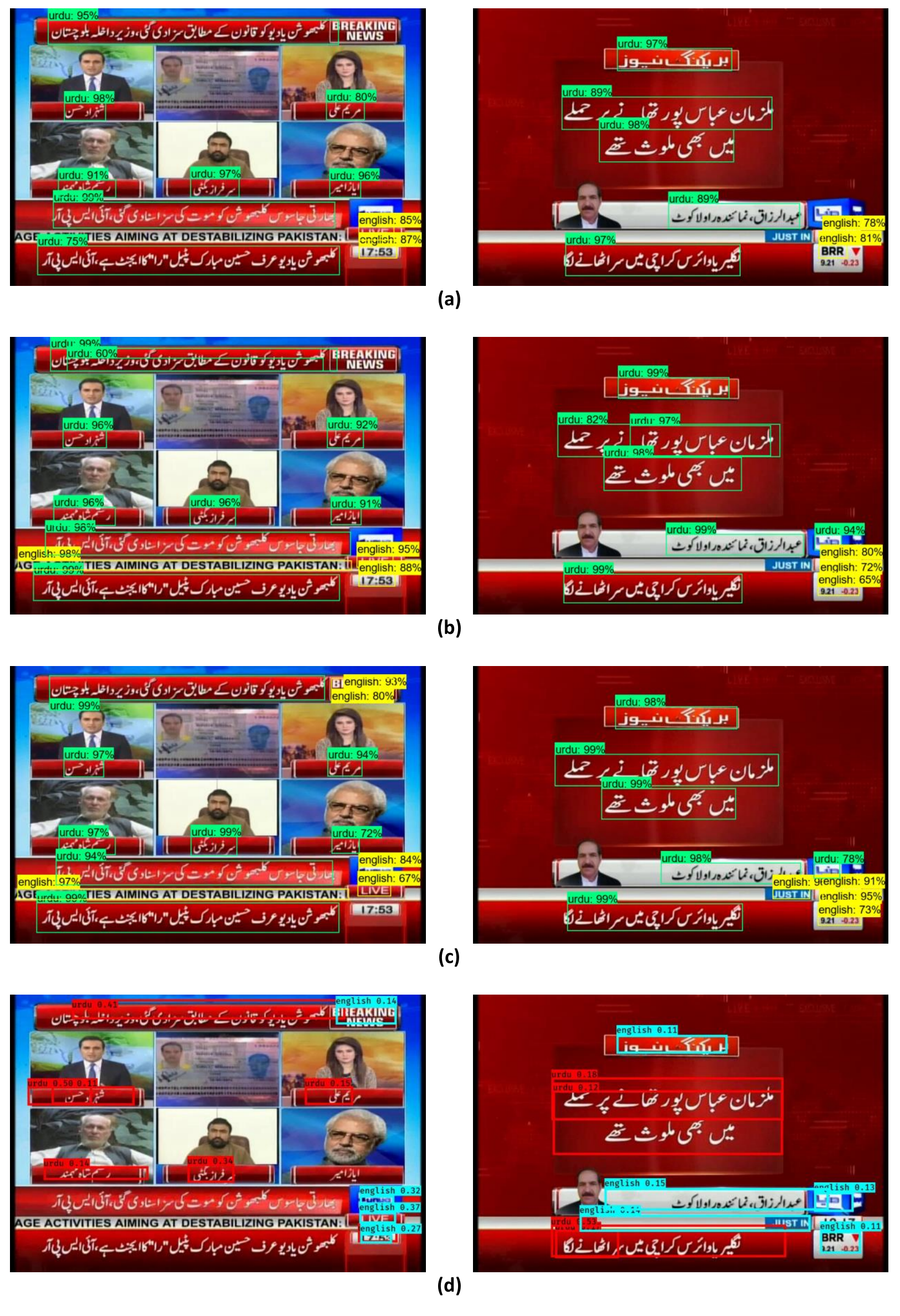}
	\caption{Detection output of hybrid text detection and script identification for different detectors (a): SSD (b): R-FCN (c): Faster RCNN (d): Yolo }
\label{fig:alldetectors}
\end{figure*}

As discussed previously, text detection and script identification can be combined in a single model treating detection as a three (rather than two) class problem. The results of these experiments are summarized in Table~\ref{tab:detectRes} keeping the same distribution of training and test frames as in the previous experiments. Many interesting observations can be drawn from the results in Table~\ref{tab:detectRes}. Similar to the script independent detectors, models pre-trained on Inception outperform those trained on ResNet and the observation is consistent for all four detectors. Likewise, Faster R-CNN reports the highest F-measure both for detection of Urdu and English text reading 0.91 and 0.87 respectively. In all cases, the performances on detection of Urdu text are better than those on Engish text. This can be attributed to the fact that the data is collected primarily from Urdu News channels which have limited amount of English text. It is also interesting to note that by combining text detection and script identification in a single model, not only the cascaded solution is avoided, the detection F-measures have also improved (in most cases). Though the improvement is marginal, eliminating the separate processing of detected text regions to identify the script offers a much simplified (yet effective) solution. Detection outputs on sample frames for the four detectors are illustrated in Figure~\ref{fig:alldetectors}.

\begin{table*}[!ht]
\centering
\caption{Performance of hybrid text detector and script identifier}
\label{tab:detectRes}
%\scriptsize
%\footnotesize
\begin{tabular}{llcccccc}
\hline%\cline{3-8}
                                                             &                  & \multicolumn{3}{c}{\textbf{RestNet}}                     & \multicolumn{3}{c}{\textbf{Inception}}                   \\ \hline
\multicolumn{1}{l}{\textbf{Method}}                         & \textbf{Script}  & \textbf{Precision} & \textbf{Recall} & \textbf{F-Measure} & \textbf{Precision} & \textbf{Recall} & \textbf{F-Measure} \\ \hline
\multicolumn{1}{l}{\multirow{3}{*}{\textbf{SSD}}}          & \textbf{Urdu}    & 0.83               & 0.72            & 0.77               & 0.82               & 0.78            & 0.80               \\ %\cline{2-8} 
\multicolumn{1}{l}{}                                       & \textbf{English} & 0.80               & 0.63            & 0.70               & 0.82               & 0.70            & 0.75               \\ %\cline{2-8} 
\hline
\multicolumn{1}{l}{\multirow{3}{*}{\textbf{R-FCN}}}        & \textbf{Urdu}    & 0.80               & 0.87            & 0.83               & 0.85               & 0.90            & 0.87               \\ %\cline{2-8} 
\multicolumn{1}{l}{}                                       & \textbf{English} & 0.73               & 0.81            & 0.77               & 0.77               & 0.84            & 0.81               \\ %\cline{2-8} 
\hline
\multicolumn{1}{l}{\multirow{3}{*}{\textbf{Faster R-CNN}}} & \textbf{Urdu}    & 0.82               & 0.92            & 0.86               & \textbf{0.87}      & \textbf{0.95}   & \textbf{0.91}      \\ %\cline{2-8} 
\multicolumn{1}{l}{}                                       & \textbf{English} & 0.80               & 0.81            & 0.80               & \textbf{0.81}      & \textbf{0.94}   & \textbf{0.87}      \\ %\cline{2-8} 
\hline
\multicolumn{1}{l}{\multirow{3}{*}{\textbf{Yolo}}}         & \textbf{Urdu}    & -                    &   -              &    -                & 0.64               & 0.70            & 0.67               \\ %\cline{2-8} 
\multicolumn{1}{l}{}                                       & \textbf{English} &  -                  &    -             &    -               & 0.62               & 0.67            & 0.64               \\ %\cline{2-8} 
\hline
\end{tabular}
\end{table*}

\begin{figure}[!ht]
	\centering
		\includegraphics[width=0.75\textwidth]{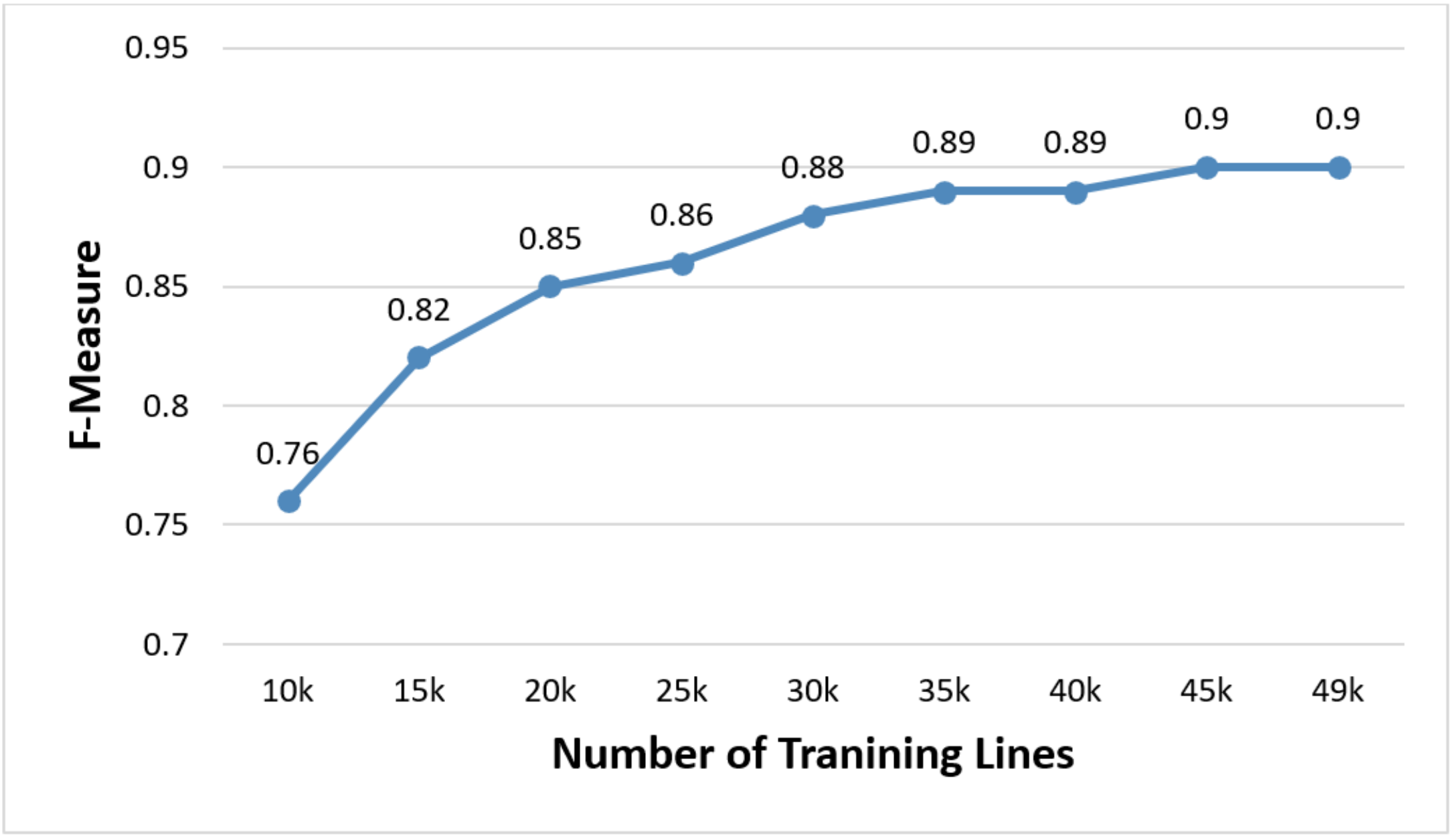}
	\caption{Impact of size of training data on text detection performance (Faster R-CNN with Inception)}
\label{fig:impact}
\end{figure}

In an attempt to carry out an in-depth analysis of the detection performance and its evolution with respect to important system parameters, we carried out another series of experiments using Faster R-CNN (with Inception). In the first such experiment, we study the performance sensitivity to the amount of training data. We train the model by varying the number of text line images (from 10K to 49K) and compute the detector F-measure. Naturally, the detector performance enhances with the increase in the amount of training data (Figure~\ref{fig:impact}) and begins to stabilize from around 30K-35K training lines. 

\begin{figure}[!ht]
	\centering
		\includegraphics[width=0.75\textwidth]{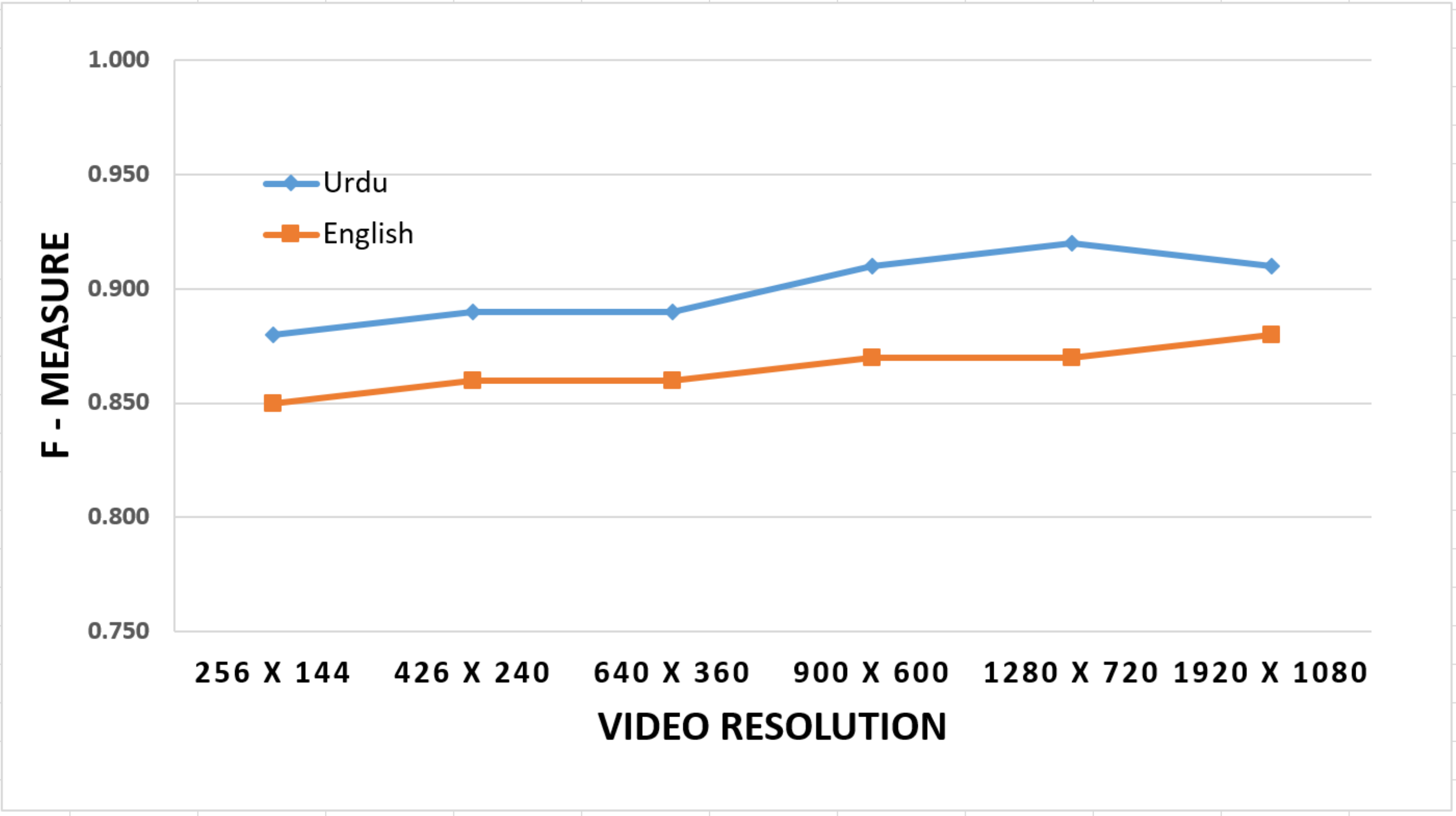}
	\caption{Impact of video resolution on text detection performance (Faster R-CNN with Inception)}
\label{fig:resolution}
\end{figure}

Resolution of input video frames is an important parameter that might affect the detector performance. To study the detector sensitivity to image resolution, we varied the image resolution from $256 \times 144$ to $1920 \times 1080$. The resolution was varied only in the test set and all sets of images were evaluated on the detector trained on a single resolution ($900 \times 600$). The F-measures in Figure~\ref{fig:resolution} are more less consistent for varied image resolutions reflecting the robustness of the detector. The proposed anchor boxes adapted for textual content play a key role in achieving this scale invariance.

%We also separately compute the detection F-measures on each of the four News channels considered in our study. The detection performance on Urdu text seems to be consistent across the four channels, for English text however, the performances are relatively low and depict more variation (Figure~\ref{fig:channels}). As discussed previously, this can be attributed to the relatively lesser number of English text lines in the training set and their varied distribution in the various channels. 
%\begin{figure}[!ht]
%	\centering		\includegraphics[width=0.5\textwidth]{Figure17.png}
%	\caption{Text Detection F-Measures on four different News channels}
%\label{fig:channels}
%\end{figure}

\subsection{Performance Comparison}
In an attempt to compare the performance of our detector with those reported in the literature, we present a comparative overview of various text detectors targeting cursive caption text in Table~\ref{tab:detectResCom}. It is important to note that since different studies are evaluated on different datasets, a direct comparison of these techniques is difficult. Most of the listed studies employ a small set of images ($\leq 1000$). Moradi et al.~\cite{moradi2013hybrid} and Zayene et al.~\cite{Oussama2016} report results on relatively larger datasets with F-measures of 0.89 and 0.84 respectively. In comparison to other studies, we employ a significantly larger set of images with an F-measure of 0.91. Furthermore, for a fair comparison, we also evaluated our system on the set 1000 images in the publicly available IPC dataset~\cite{siddiqi2012database}, the corresponding F-measure reads 0.92 validating the effectiveness of our detection technique,

\begin{table*}[!ht]
\centering
\caption{Performance comparison with other techniques}
\label{tab:detectResCom}
\footnotesize
%\scriptsize
\begin{tabular}{llllcccc}
\hline
\multicolumn{1}{l}{\textbf{Study}}                              & 
\multicolumn{1}{l}{\textbf{Method}} &
\multicolumn{1}{l}{\textbf{Dataset}}                              & 
\multicolumn{1}{l}{\textbf{Script}}   & \multicolumn{1}{c}{\textbf{Video Frames}} & \multicolumn{1}{c}{\textbf{Precision}} & \multicolumn{1}{c}{\textbf{Recall}} & \multicolumn{1}{c}{\textbf{F-Measure}} \\ \hline

Jamil et al.(2011)~\cite{jamil2011edge}   & Edge-based Features   & IPC & Urdu    & 150    & 0.77  & 0.81      & 0.79          \\
Siddiqi and Raza(2012)~\cite{siddiqi2012database} & Image Analysis & IPC & Urdu    & 1,000   & 0.71  & 0.80     & 0.75            \\
Moradi et al.(2013)~\cite{moradi2013hybrid}   & LBP with SVM & -    & Farsi/Arabic & 4971 & 0.91 & 0.87   & 0.89          \\
Raza et al.(2013)~\cite{raza2013multilingual} & Cascade of Transforms  &  IPC  & Urdu    & 1,000   & 0.80      & 0.89   & 0.84       \\
Raza et al.(2013)~\cite{raza2013multilingual}& Cascade of Transforms  &  IPC    & Arabic & 300 & 0.81  & 0.93    & 0.86      \\
Yousfi et al.(2014)~\cite{yousfi2014arabic}   & ConvNet    & - & Arabic & 201     & 0.75    & 0.80     & 0.77                  \\
Zayene et al.(2015)~\cite{Oussama2015}    & SWT & AcTiV        & Arabic  & 425     & 0.67   & 0.73     & 0.70                   \\
Zayene et al.(2016)~\cite{Oussama2016}    & SWT\&Conv Autoencoders  & AcTiV  & Arabic   & 1843    & 0.83      & 0.85     & 0.84        \\
Shahzad et al.(2017)~\cite{shahzad2017oriental}  & Image Analysis & - & Urdu/Arabic & 240 & 0.83   & 0.93   & 0.88                  \\
Mirza et al.(2018)~\cite{mirza2018urdu}  & Textural Features & UTiV & Urdu  & 1,000    & 0.72   & 0.89    & 0.80           \\
Unar et al.(2018)~\cite{unar2018artificial}  & Image Analysis+SVM & IPC     & Urdu   & 1,000  & 0.83    & 0.88     & 0.85            \\
\textbf{Proposed Method}    & \textbf{Deep ConvNets} & UTiV  & \textbf{Urdu}&    \textbf{11,203} & \textbf{0.87}  & \textbf{0.95} & \textbf{0.91}\\ 
& & IPC  & \textbf{Urdu} &    \textbf{1,000} & \textbf{0.91}  & \textbf{0.93} & \textbf{0.92}\\ 

\hline     

\end{tabular}
\end{table*}

\section{Conclusion}\label{sec:Conclusion}
This paper presented a system for detection of caption text appearing in video frames. The developed technique relies on exploiting deep learning based object detectors and adapting them for text detection. Since it is common in videos to have text in more than one script, we presented, as a case study, video frames with text in cursive (Urdu) and Roman (English) scripts. Since each script requires different processing, the detection is combined with script identification in an end-to-end fashion so that the system is able to not only localize the text but also identify its script. Among various investigated object detectors, Faster R-CNN with our proposed set of anchor boxes reported the highest detection rates.\\

The presented work is a part of a comprehensive video indexing and retrieval system and the current study focused on the detection of text. In our other~\cite{ali1,ali2} work, the detection module is  integrated with the video OCR module so that detected text is recognized. Once recognized, videos are indexed based on keywords and retrieved on user provided queries. In addition to retrieval, automatic News summarization from ticker text and comparison of News across various News channels is also planned to be implemented. Furthermore, the textual content based retrieval will be combined with audio as well as visual objects appearing in videos.

%\clearpage
%\bibliographystyle{plain} 	
%\bibliography{References}

\begin{thebibliography}{10}

\bibitem{anthimopoulos2010two}
Marios Anthimopoulos, Basilis Gatos, and Ioannis Pratikakis.
\newblock A two-stage scheme for text detection in video images.
\newblock {\em Image and Vision Computing}, 28(9):1413--1426, 2010.

\bibitem{XiangBai2016}
Xiang Bai, Cong Yao, and Wenyu Liu.
\newblock Strokelets: A learned multi-scale mid-level representation for scene
  text recognition.
\newblock {\em IEEE TRANSACTIONS ON IMAGE PROCESSING}, 25, 2016.

\bibitem{banerjee2013robust}
Sudipto Banerjee, Koustav Mullick, and Ujjwal Bhattacharya.
\newblock A robust approach to extraction of texts from camera captured images.
\newblock In {\em International Workshop on Camera-Based Document Analysis and
  Recognition}, pages 30--46. Springer, 2013.

\bibitem{baran2018automated}
Remigiusz Baran, Pavol Partila, and Rafal Wilk.
\newblock Automated text detection and character recognition in natural scenes
  based on local image features and contour processing techniques.
\newblock In {\em International Conference on Intelligent Human Systems
  Integration}, pages 42--48. Springer, 2018.

\bibitem{bouchain2006character}
David Bouchain.
\newblock Character recognition using convolutional neural networks.
\newblock {\em Institute for Neural Information Processing}, 2007, 2006.

\bibitem{burgess2018youtube}
Jean Burgess and Joshua Green.
\newblock {\em YouTube: Online video and participatory culture}.
\newblock John Wiley \& Sons, 2018.

\bibitem{dai2016r}
Jifeng Dai, Yi~Li, Kaiming He, and Jian Sun.
\newblock R-fcn: Object detection via region-based fully convolutional
  networks.
\newblock In {\em Advances in neural information processing systems}, pages
  379--387, 2016.

\bibitem{dai2018scene}
Jin Dai, Zu~Wang, Xianjing Zhao, and Shuai Shao.
\newblock Scene text detection based on enhanced multi-channels mser and a fast
  text grouping process.
\newblock In {\em 2018 IEEE 3rd International Conference on Cloud Computing and
  Big Data Analysis (ICCCBDA)}, pages 351--355. IEEE, 2018.

\bibitem{dalal2005histograms}
Navneet Dalal and Bill Triggs.
\newblock Histograms of oriented gradients for human detection.
\newblock In {\em Computer Vision and Pattern Recognition, 2005. CVPR 2005.
  IEEE Computer Society Conference on}, volume~1, pages 886--893. IEEE, 2005.

\bibitem{farfade2015multi}
Sachin~Sudhakar Farfade, Mohammad~J Saberian, and Li-Jia Li.
\newblock Multi-view face detection using deep convolutional neural networks.
\newblock In {\em Proceedings of the 5th ACM on International Conference on
  Multimedia Retrieval}, pages 643--650. ACM, 2015.

\bibitem{gabor1946}
Dennis Gabor.
\newblock Theory of communication. part 1: The analysis of information.
\newblock {\em Journal of the Institution of Electrical Engineers-Part III:
  Radio and Communication Engineering}, 93(26):429--441, 1946.

\bibitem{girshick2015fast}
Ross Girshick.
\newblock Fast r-cnn.
\newblock In {\em Proceedings of the IEEE international conference on computer
  vision}, pages 1440--1448, 2015.

\bibitem{girshick2014rich}
Ross Girshick, Jeff Donahue, Trevor Darrell, and Jitendra Malik.
\newblock Rich feature hierarchies for accurate object detection and semantic
  segmentation.
\newblock In {\em Proceedings of the IEEE conference on computer vision and
  pattern recognition}, pages 580--587, 2014.

\bibitem{joutel2007curvelets}
Joutel Guillaume, Eglin V{\'e}ronique, Bres St{\'e}phane, and Emptoz Hubert.
\newblock Curvelets based feature extraction of handwritten shapes for ancient
  manuscripts classification.
\newblock In {\em Electronic Imaging 2007}, pages 65000D--65000D. International
  Society for Optics and Photonics, 2007.

\bibitem{gupta2016synthetic}
Ankush Gupta, Andrea Vedaldi, and Andrew Zisserman.
\newblock Synthetic data for text localisation in natural images.
\newblock In {\em Proceedings of the IEEE Conference on Computer Vision and
  Pattern Recognition}, pages 2315--2324, 2016.

\bibitem{hayat2018ligature}
Umar Hayat, Muhammad Aatif, Osama Zeeshan, and Imran Siddiqi.
\newblock Ligature recognition in urdu caption text using deep convolutional
  neural networks.
\newblock In {\em 2018 14th International Conference on Emerging Technologies
  (ICET)}, pages 1--6. IEEE, 2018.

\bibitem{he2016deep}
Kaiming He, Xiangyu Zhang, Shaoqing Ren, and Jian Sun.
\newblock Deep residual learning for image recognition.
\newblock In {\em Proceedings of the IEEE conference on computer vision and
  pattern recognition}, pages 770--778, 2016.

\bibitem{he2018single}
Tong He, Zhi Tian, Weilin Huang, Chunhua Shen, Yu~Qiao, and Changming Sun.
\newblock Single shot textspotter with explicit alignment and attention.
\newblock {\em arXiv preprint arXiv:1803.03474}, 2018.

\bibitem{huang2017speed}
Jonathan Huang, Vivek Rathod, Chen Sun, Menglong Zhu, Anoop Korattikara,
  Alireza Fathi, Ian Fischer, Zbigniew Wojna, Yang Song, Sergio Guadarrama,
  et~al.
\newblock Speed/accuracy trade-offs for modern convolutional object detectors.
\newblock In {\em Proceedings of the IEEE conference on computer vision and
  pattern recognition}, pages 7310--7311, 2017.

\bibitem{huang2013scene}
Rong Huang, Palaiahnakote Shivakumara, and Seiichi Uchida.
\newblock Scene character detection by an edge-ray filter.
\newblock In {\em Document Analysis and Recognition (ICDAR), 2013 12th
  International Conference on}, pages 462--466. IEEE, 2013.

\bibitem{huang2014robust}
Weilin Huang, Yu~Qiao, and Xiaoou Tang.
\newblock Robust scene text detection with convolution neural network induced
  mser trees.
\newblock In {\em European Conference on Computer Vision}, pages 497--511.
  Springer, 2014.

\bibitem{Huang:2011}
X~Huang.
\newblock A novel video text extraction approach based on log-gabor filters.
\newblock In {\em 4th International Congress on Image and Signal Processing},
  2011.

\bibitem{ioffe2015batch}
Sergey Ioffe and Christian Szegedy.
\newblock Batch normalization: Accelerating deep network training by reducing
  internal covariate shift.
\newblock {\em arXiv preprint arXiv:1502.03167}, 2015.

\bibitem{Jamil:2011}
A~Jamil, I~Siddiqi, F~Arif, and A~Raza.
\newblock Edge-based features for localization of artificial urdu text in video
  images.
\newblock In {\em International Conference on Document Analysis and
  Recognition}, Jamil, A., et al. Edge-based Features for Localization of
  Artificial Urdu Text in Video Images. in International Conference on 2011.

\bibitem{ali2016}
Akhtar Jamil, Azra Batool, Zumra Malik, Ali Mirza, and Imran Siddiqi.
\newblock Multilingual artificial text extraction and script identification
  from video images.
\newblock {\em International Journal of Advanced Computer Science \&
  Applications}, 1(7):529--539, 2016.

\bibitem{jamil2011edge}
Akhtar Jamil, Imran Siddiqi, Fahim Arif, and Ahsen Raza.
\newblock Edge-based features for localization of artificial urdu text in video
  images.
\newblock In {\em Document Analysis and Recognition (ICDAR), 2011 International
  Conference on}, pages 1120--1124. IEEE, 2011.

\bibitem{Kiran:2012}
Y.C. Kiran and L.N. C.
\newblock Text extraction and verification from video based on svm.
\newblock {\em World Journal of Science and Technology}, 2(5):124--126, 2012.

\bibitem{koo2013scene}
Hyung~Il Koo and Duck~Hoon Kim.
\newblock Scene text detection via connected component clustering and nontext
  filtering.
\newblock {\em IEEE transactions on image processing}, 22(6):2296--2305, 2013.

\bibitem{krizhevsky2012imagenet}
Alex Krizhevsky, Ilya Sutskever, and Geoffrey~E Hinton.
\newblock Imagenet classification with deep convolutional neural networks.
\newblock In {\em Advances in neural information processing systems}, pages
  1097--1105, 2012.

\bibitem{lecun2015deep}
Yann LeCun, Yoshua Bengio, and Geoffrey Hinton.
\newblock Deep learning.
\newblock {\em nature}, 521(7553):436, 2015.

\bibitem{lee2010scene}
SeongHun Lee, Min~Su Cho, Kyomin Jung, and Jin~Hyung Kim.
\newblock Scene text extraction with edge constraint and text collinearity.
\newblock In {\em Pattern Recognition (ICPR), 2010 20th International
  Conference on}, pages 3983--3986. IEEE, 2010.

\bibitem{liao2017textboxes}
Minghui Liao, Baoguang Shi, Xiang Bai, Xinggang Wang, and Wenyu Liu.
\newblock Textboxes: A fast text detector with a single deep neural network.
\newblock In {\em AAAI}, pages 4161--4167, 2017.

\bibitem{lin2014microsoft}
Tsung-Yi Lin, Michael Maire, Serge Belongie, James Hays, Pietro Perona, Deva
  Ramanan, Piotr Doll{\'a}r, and C~Lawrence Zitnick.
\newblock Microsoft coco: Common objects in context.
\newblock In {\em European conference on computer vision}, pages 740--755.
  Springer, 2014.

\bibitem{liu2016ssd}
Wei Liu, Dragomir Anguelov, Dumitru Erhan, Christian Szegedy, Scott Reed,
  Cheng-Yang Fu, and Alexander~C Berg.
\newblock Ssd: Single shot multibox detector.
\newblock In {\em European conference on computer vision}, pages 21--37.
  Springer, 2016.

\bibitem{lucas2005icdar}
Simon~M Lucas, Alex Panaretos, Luis Sosa, Anthony Tang, Shirley Wong, Robert
  Young, Kazuki Ashida, Hiroki Nagai, Masayuki Okamoto, Hiroaki Yamamoto,
  et~al.
\newblock Icdar 2003 robust reading competitions: entries, results, and future
  directions.
\newblock {\em International Journal of Document Analysis and Recognition
  (IJDAR)}, 7(2-3):105--122, 2005.

\bibitem{mirza2018urdu}
Ali Mirza, Marium Fayyaz, Zunera Seher, and Imran Siddiqi.
\newblock Urdu caption text detection using textural features.
\newblock In {\em Proceedings of the 2nd Mediterranean Conference on Pattern
  Recognition and Artificial Intelligence}, pages 70--75. ACM, 2018.

\bibitem{ali2019}
Ali Mirza, Imran Siddiqi, Syed G.Mustufa, and Mazahir Hussain.
\newblock Impact of pre-processing on recognition of cursive video text.
\newblock In {\em 9th Iberian Conference on Pattern Recognition and Image
  Analysis}. Springer, 2019.

\bibitem{moradi2013hybrid}
Mohieddin Moradi and Saeed Mozaffari.
\newblock Hybrid approach for farsi/arabic text detection and localisation in
  video frames.
\newblock {\em IET Image Processing}, 7(2):154--164, 2013.

\bibitem{pan2011hybrid}
Yi-Feng Pan, Xinwen Hou, and Cheng-Lin Liu.
\newblock A hybrid approach to detect and localize texts in natural scene
  images.
\newblock {\em IEEE Transactions on Image Processing}, 20(3):800--813, 2011.

\bibitem{Ye2015}
Ye~Qiaoyang and David Doermann.
\newblock Text detection and recognition in imagery: A survey.
\newblock {\em IEEE Transactions On Pattern Analysis And Machine Intelligence},
  37, July 2015.

\bibitem{raza2013multilingual}
Ahsen Raza, Imran Siddiqi, Chawki Djeddi, and Abdellatif Ennaji.
\newblock Multilingual artificial text detection using a cascade of transforms.
\newblock In {\em 2013 12th International Conference on Document Analysis and
  Recognition}, pages 309--313. IEEE, 2013.

\bibitem{redmon2016you}
Joseph Redmon, Santosh Divvala, Ross Girshick, and Ali Farhadi.
\newblock You only look once: Unified, real-time object detection.
\newblock In {\em Proceedings of the IEEE conference on computer vision and
  pattern recognition}, pages 779--788, 2016.

\bibitem{redmon2017yolo9000}
Joseph Redmon and Ali Farhadi.
\newblock Yolo9000: better, faster, stronger.
\newblock In {\em Proceedings of the IEEE conference on computer vision and
  pattern recognition}, pages 7263--7271, 2017.

\bibitem{redmon2018yolov3}
Joseph Redmon and Ali Farhadi.
\newblock Yolov3: An incremental improvement.
\newblock {\em arXiv preprint arXiv:1804.02767}, 2018.

\bibitem{ren2015faster}
Shaoqing Ren, Kaiming He, Ross Girshick, and Jian Sun.
\newblock Faster r-cnn: Towards real-time object detection with region proposal
  networks.
\newblock In {\em Advances in neural information processing systems}, pages
  91--99, 2015.

\bibitem{russakovsky2015imagenet}
Olga Russakovsky, Jia Deng, Hao Su, Jonathan Krause, Sanjeev Satheesh, Sean Ma,
  Zhiheng Huang, Andrej Karpathy, Aditya Khosla, Michael Bernstein, et~al.
\newblock Imagenet large scale visual recognition challenge.
\newblock {\em International Journal of Computer Vision}, 115(3):211--252,
  2015.

\bibitem{shahzad2017oriental}
Usman Shahzad and Khurram Khurshid.
\newblock Oriental-script text detection and extraction in videos.
\newblock In {\em 2017 1st International Workshop on Arabic Script Analysis and
  Recognition (ASAR)}, pages 15--20. IEEE, 2017.

\bibitem{sharma2012recent}
Nabin Sharma, Umapada Pal, and Michael Blumenstein.
\newblock Recent advances in video based document processing: a review.
\newblock In {\em Document Analysis Systems (DAS), 2012 10th IAPR International
  Workshop on}, pages 63--68. IEEE, 2012.

\bibitem{shi2017detecting}
Baoguang Shi, Xiang Bai, and Serge Belongie.
\newblock Detecting oriented text in natural images by linking segments.
\newblock {\em arXiv preprint arXiv:1703.06520}, 2017.

\bibitem{Shivakumara:2010}
P.~Shivakumara, T.Q. Phan, and C.L. Tan.
\newblock New fourier-statistical features in rgb space for video text
  detection.
\newblock {\em IEEE Trans. Circuits Syst. Video Techn}, pages 1520--1532, 2010.

\bibitem{shivakumara2010new}
Palaiahnakote Shivakumara, Trung~Quy Phan, and Chew~Lim Tan.
\newblock New fourier-statistical features in rgb space for video text
  detection.
\newblock {\em IEEE Transactions on Circuits and Systems for Video Technology},
  20(11):1520--1532, 2010.

\bibitem{shivakumara2012multioriented}
Palaiahnakote Shivakumara, Rushi~Padhuman Sreedhar, Trung~Quy Phan, Shijian Lu,
  and Chew~Lim Tan.
\newblock Multioriented video scene text detection through bayesian
  classification and boundary growing.
\newblock {\em IEEE Transactions on Circuits and systems for Video Technology},
  22(8):1227--1235, 2012.

\bibitem{shrivastava2017learning}
Ashish Shrivastava, Tomas Pfister, Oncel Tuzel, Joshua Susskind, Wenda Wang,
  and Russell Webb.
\newblock Learning from simulated and unsupervised images through adversarial
  training.
\newblock In {\em CVPR}, volume~2, page~5, 2017.

\bibitem{siddiqi2012database}
Imran Siddiqi and Ahsen Raza.
\newblock A database of artificial urdu text in video images with
  semi-automatic text line labeling scheme.
\newblock In {\em MMEDIA 2012, The Fourth International Conferences on Advances
  in Multimedia}, pages 75--81, 2012.

\bibitem{simonyan2014very}
Karen Simonyan and Andrew Zisserman.
\newblock Very deep convolutional networks for large-scale image recognition.
\newblock {\em arXiv preprint arXiv:1409.1556}, 2014.

\bibitem{szegedy2015going}
Christian Szegedy, Wei Liu, Yangqing Jia, Pierre Sermanet, Scott Reed, Dragomir
  Anguelov, Dumitru Erhan, Vincent Vanhoucke, and Andrew Rabinovich.
\newblock Going deeper with convolutions.
\newblock In {\em Proceedings of the IEEE conference on computer vision and
  pattern recognition}, pages 1--9, 2015.

\bibitem{tian2016detecting}
Zhi Tian, Weilin Huang, Tong He, Pan He, and Yu~Qiao.
\newblock Detecting text in natural image with connectionist text proposal
  network.
\newblock In {\em European conference on computer vision}, pages 56--72.
  Springer, 2016.

\bibitem{uijlings2013selective}
Jasper~RR Uijlings, Koen~EA Van De~Sande, Theo Gevers, and Arnold~WM Smeulders.
\newblock Selective search for object recognition.
\newblock {\em International journal of computer vision}, 104(2):154--171,
  2013.

\bibitem{unar2018artificial}
Salahuddin Unar, Akhtar~Hussain Jalbani, Muhammad~Moazzam Jawaid, Mohsin
  Shaikh, and Asghar~Ali Chandio.
\newblock Artificial urdu text detection and localization from individual video
  frames.
\newblock {\em Mehran University Research Journal of Engineering and
  Technology}, 37(2):429--438, 2018.

\bibitem{wang2016}
Shupeng Wang, Chenglin Fu, and Qi~Li.
\newblock Text detection in natural scene image: A survey.
\newblock In {\em International Conference on Machine Learning and Intelligent
  Communications}, pages 257--264. Springer, 2016.

\bibitem{wang2018crf}
Yanna Wang, Cunzhao Shi, Baihua Xiao, Chunheng Wang, and Chengzuo Qi.
\newblock Crf based text detection for natural scene images using convolutional
  neural network and context information.
\newblock {\em Neurocomputing}, 295:46--58, 2018.

\bibitem{wolf2006object}
Christian Wolf and Jean-Michel Jolion.
\newblock Object count/area graphs for the evaluation of object detection and
  segmentation algorithms.
\newblock {\em International Journal of Document Analysis and Recognition
  (IJDAR)}, 8(4):280--296, 2006.

\bibitem{xu2018end}
Xu~Yan, Shan Siyuan, Qiu Ziming, Jia Zhipeng, Shen Zhengyang, Wang Yipei, Shi
  Mengfei, Eric I, and Chang Chao.
\newblock End-to-end subtitle detection and recognition for videos in east
  asian languages via cnn ensemble.
\newblock {\em Signal Processing: Image Communication}, 60:131--143, 2018.

\bibitem{ye2005fast}
Qixiang Ye, Qingming Huang, Wen Gao, and Debin Zhao.
\newblock Fast and robust text detection in images and video frames.
\newblock {\em Image and Vision Computing}, 23(6):565--576, 2005.

\bibitem{yi2012localizing}
Chucai Yi and Yingli Tian.
\newblock Localizing text in scene images by boundary clustering, stroke
  segmentation, and string fragment classification.
\newblock {\em IEEE Transactions on Image Processing}, 21(9):4256--4268, 2012.

\bibitem{Yi:2007}
J~Yi, Y~Peng, and J~Xiao.
\newblock Color-based clustering for text detection and extraction in image.
\newblock In {\em 15th international conference on Multimedia, Germany.}, 2007.

\bibitem{Yin:2013}
X.-C.and X.~Yin Yin and K.~Huang.
\newblock Robust text detection in natural scene images.
\newblock {\em CoRR abs/1301.2628.}, 2013.

\bibitem{yin2016text}
Xu-Cheng Yin, Ze-Yu Zuo, Shu Tian, and Cheng-Lin Liu.
\newblock Text detection, tracking and recognition in video: A comprehensive
  survey.
\newblock {\em IEEE Transactions on Image Processing}, 25(6):2752--2773, 2016.

\bibitem{yousfi2014arabic}
Sonia Yousfi, Sid-Ahmed Berrani, and Christophe Garcia.
\newblock Arabic text detection in videos using neural and boosting-based
  approaches: Application to video indexing.
\newblock In {\em 2014 IEEE International Conference on Image Processing
  (ICIP)}, pages 3028--3032. IEEE, 2014.

\bibitem{Oussama2016}
Oussama Zayene, Jean Hennebert, Mathias Seuret, Sameh~M. Touj, Rolf Ingold, and
  Najoua Essoukri~Ben Amara.
\newblock Text detection in arabic news video based on swt operator and
  convolutional auto-encoders.
\newblock {\em 2016 12th IAPR Workshop on Document Analysis Systems}, 2016.

\bibitem{Oussama2015}
Oussama Zayene, Jean Hennebert, Sameh~Masmoudi Touj, Rolf Ingold, and Najoua
  Essoukri~Ben Amara.
\newblock A dataset for arabic text detection, tracking and recognition in news
  videos - activ.
\newblock {\em 2015 13th International Conference on Document Analysis and
  Recognition (ICDAR)}, 2015.

\bibitem{zhang2013}
Honggang Zhang, Kaili Zhao, Yi-Zhe Song, and Jun Guo.
\newblock Text extraction from natural scene image: A survey.
\newblock {\em Neurocomputing}, 122:310--323, 2013.

\bibitem{zhang2016multi}
Zheng Zhang, Chengquan Zhang, Wei Shen, Cong Yao, Wenyu Liu, and Xiang Bai.
\newblock Multi-oriented text detection with fully convolutional networks.
\newblock In {\em Proceedings of the IEEE Conference on Computer Vision and
  Pattern Recognition}, pages 4159--4167, 2016.

\bibitem{Zhen:2009}
W.~Zhen and W.~Zagiqiang.
\newblock A comparative study of feature selection for svm in video text
  detection.
\newblock In {\em 2nd International symposium on Computational Intelligence and
  Design p. 552 - 556.}, 2009.

\bibitem{zhong2000automatic}
Yu~Zhong, Hongjiang Zhang, and Anil~K Jain.
\newblock Automatic caption localization in compressed video.
\newblock {\em IEEE transactions on pattern analysis and machine intelligence},
  22(4):385--392, 2000.

\bibitem{zhou2017east}
Xinyu Zhou, Cong Yao, He~Wen, Yuzhi Wang, Shuchang Zhou, Weiran He, and Jiajun
  Liang.
\newblock East: an efficient and accurate scene text detector.
\newblock In {\em Proc. CVPR}, pages 2642--2651, 2017.

\bibitem{ali1}
Ali Mirza, Ossama Zeshan, Muhammad Atif, Imran Siddiqi.
\newblock Detection and recognition of cursive text from video frames.
\newblock In {\em EURASIP Journal on Image and Video Processing}, pages 1--19, 2020.

\bibitem{ali2}
Ali Mirza, Imran Siddiqi.
\newblock Recognition of cursive video text using a deep learning framework.
\newblock In {\em IET Image Processing}, pages 3444--3455, 2020.

\end{thebibliography}

\vfill\pagebreak
\end{document}